\documentclass[11pt]{article}
\usepackage[pageonlyfooter]{evolvingmachines}
\usepackage{amsmath,amsfonts,bm}

\def\eqref#1{equation~\ref{#1}}

\def\1{\bm{1}}

\DeclareMathAlphabet{\mathsfit}{\encodingdefault}{\sfdefault}{m}{sl}
\SetMathAlphabet{\mathsfit}{bold}{\encodingdefault}{\sfdefault}{bx}{n}

\usepackage{amsmath,amssymb}
\usepackage{booktabs}
\usepackage{graphicx}
\usepackage{makecell}
\usepackage{multirow}
\usepackage{tabularx}
\usepackage{array}
\usepackage[table]{xcolor}
\usepackage{hyperref}
\usepackage{url}
\usepackage{fontawesome5}

\emltitlesize{22}{25}
\title{Arm-wise Compositional Generalization in\\
Dual-Arm Vision-Language-Action Models}

\author{%
Zaibin~Zhang\textsuperscript{1,*,\textdagger},\quad Binghao~Ran\textsuperscript{1,*},\quad
Yuhan~Wu\textsuperscript{1,*},\quad Zhongbo~Zhang\textsuperscript{1,*},\quad
Yifan~Wang\textsuperscript{1},\quad Junwei~Jiang\textsuperscript{1}\\
Junlan~Xiao\textsuperscript{1},\quad Wangcheng~Shi\textsuperscript{1},\quad
Li~Kang\textsuperscript{3},\quad Yiran~Qin\textsuperscript{4},\quad
Zhenfei~Yin\textsuperscript{2},\quad Lijun~Wang\textsuperscript{1},\quad
Huchuan~Lu\textsuperscript{1}}
\emlaffiliation{%
\makebox[\linewidth][c]{%
\textsuperscript{1}Dalian University of Technology\hspace{5mm}%
\textsuperscript{2}University of Oxford\hspace{5mm}%
\textsuperscript{3}Shanghai Jiao Tong University}\\
\makebox[\linewidth][c]{%
\textsuperscript{4}The Chinese University of Hong Kong, Shenzhen\hspace{5mm}%
\textsuperscript{5}Evolving Machines Lab}\\[1pt]
\makebox[\linewidth][c]{%
\footnotesize\textsuperscript{*}Equal contribution.\hspace{5mm}%
\textsuperscript{\textdagger}Project leader.}}
\emlcorrespondence{%
\makebox[\linewidth][c]{%
\href{https://github.com/zhangzaibin/future-robots}{\faGithub\enspace Code}%
\hspace{6mm}%
\href{https://zhangzaibin.github.io/evolving-machines/research/arm-wise-cg/}{\faGlobe\enspace Project Page}%
\hspace{6mm}%
\href{mailto:dlutzzb@gmail.com}{\faEnvelope\enspace dlutzzb@gmail.com}}}

\AtBeginDocument{\hypersetup{pdfauthor={Zaibin Zhang, Binghao Ran, Yuhan Wu,
Zhongbo Zhang, Yifan Wang, Junwei Jiang, Junlan Xiao, Wangcheng Shi, Li Kang,
Yiran Qin, Zhenfei Yin, Lijun Wang, Huchuan Lu}}}

\emlshorttitle{Arm-wise Compositional Generalization}
\emlkeywords{Dual-arm manipulation; compositional generalization; vision-language-action models}
\date{}
\emlheader{%
  \begin{minipage}[c]{10mm}
    \centering
    \includegraphics[height=10mm]{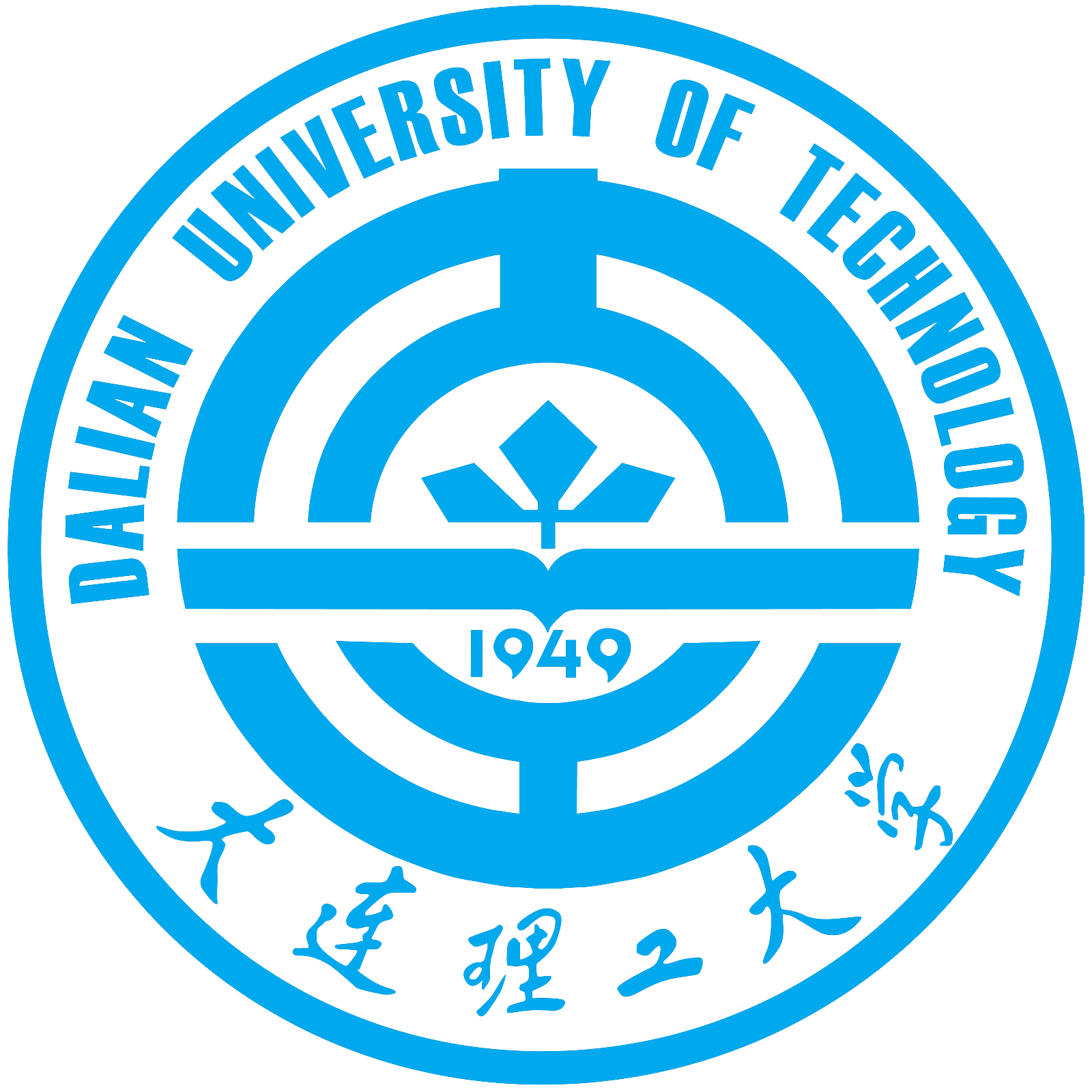}%
  \end{minipage}\hspace{5mm}%
  \begin{minipage}[c]{34mm}
    \centering
    \includegraphics[width=34mm,trim={346bp 644bp 346bp 644bp},clip]
      {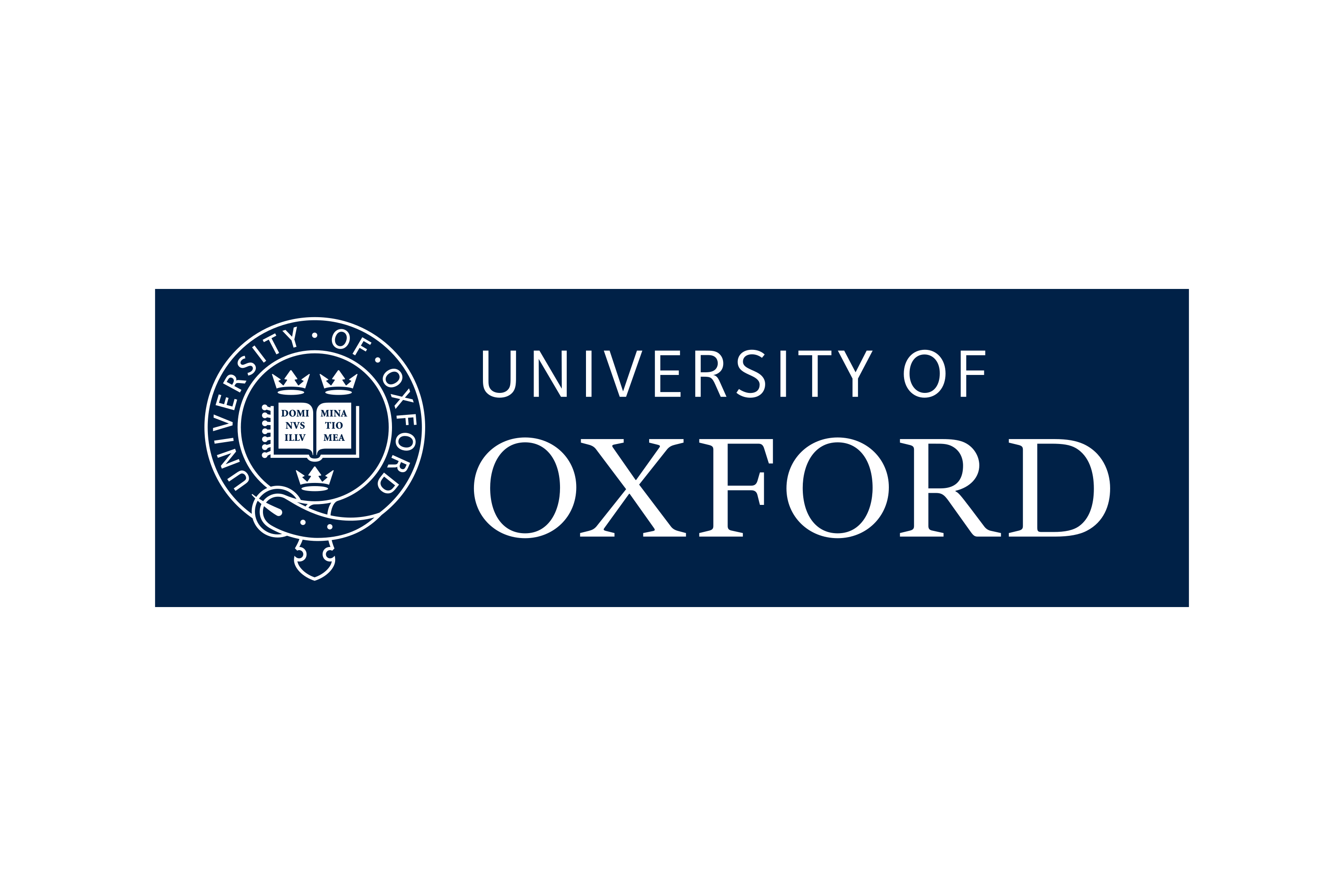}%
  \end{minipage}\hspace{5mm}%
  \begin{minipage}[c]{32mm}
    \centering
    \resizebox{32mm}{!}{%
      \begin{tabular}[c]{@{}c@{\hspace{3.5mm}}l@{}}
        \begin{tabular}[c]{@{}c@{}}\EMLMark[12mm]\end{tabular} & \EMLWordmark
      \end{tabular}}%
  \end{minipage}%
}
\begin{document}
\maketitle

\noindent\begin{minipage}{\linewidth}
    \captionsetup{type=figure}
    \centering
    \includegraphics[width=\linewidth]{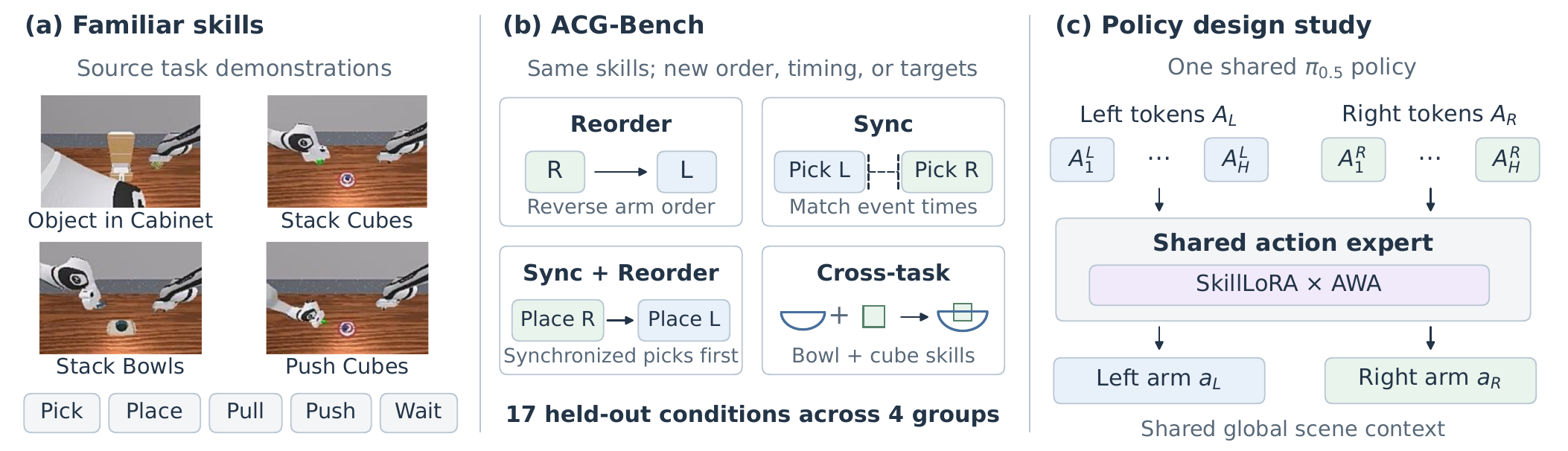}
    \caption{\textbf{ACG-Bench and the design study.} We test familiar skills
    under new order, timing, and cross-task requirements. Using one shared
    $\pi_{0.5}$ policy, we compare token grouping, skill-specific adapters,
    and arm-wise attention.}
    \label{fig:teaser}
\end{minipage}\par
\vspace{3mm}

\begin{abstract}
  Generalization in multi-arm collaboration can be studied as composing familiar atomic skills in new ways across arms. However, existing evaluations offer limited insight into which training and architectural choices support this ability under different coordination requirements. We introduce \textbf{ACG-Bench}, a benchmark for \emph{Arm-wise Compositional Generalization} that provides a common testbed for studying skill recomposition in dual-arm policies. It contains 23 task--condition pairs across 8 task families, with 6 in-domain conditions and 17 unseen compositions covering reordering, synchronization, their combination, and cross-task composition. All methods receive the same per-arm atomic prompts, and success requires achieving the task goal while satisfying physical milestones and specified order or timing constraints. Using $\pi_{0.5}$ as a common vision-language-action backbone, we compare representative data-augmentation and architectural strategies with shared source data and a common evaluation protocol. Our architectural study examines arm-token grouping, skill-specific LoRA adapters (SkillLoRA), and arm-wise attention (AWA), highlighting the complementarity of skill-conditioned parameters and attention structure. Combining these choices yields \textbf{AE-VLA}, which achieves 21.53\% generalization success in simulation, compared with 2.94\% for Single $\pi_{0.5}$, 3.06\% for MA-VLA, and 5.53\% for two independently controlled $\pi_{0.5}$ policies. On physical SO101 robots, AE-VLA reaches 39.00\% mean success across five unseen conditions, compared with 10.00\% for the strongest baseline. These findings provide empirical guidance for designing dual-arm policies that generalize beyond fixed training routines.
\end{abstract}

\clearpage
\section{Introduction}
\label{sec:intro}

Dual-arm vision-language-action (VLA) models have made progress in learning manipulation tasks~\citep{black2024pi_0,intelligence2025pi}, but executing familiar tasks does not establish whether their component skills can be recombined.
The same placing skills may need to run in a different order or synchronize their completion, while skills learned in separate tasks may need to work together.
We call this ability \emph{Arm-wise Compositional Generalization} (ACG), ie., reusing familiar arm-level skills under training-time unseen coordination requirements.
Such reuse offers a path beyond fixed collaboration routines without collecting demonstrations for every possible combination.

Existing benchmarks study skill transfer and composition through language grounding, skill chaining, and atomic-to-composite manipulation~\citep{mees2022calvin,jiang2023vima,liu2023libero,haresh2024clevrskills,wu2026atombench}, whereas dual-arm benchmarks mainly evaluate coordination patterns seen during training~\citep{mu2025robotwin,chen2025robotwin,chen2026robodojo}.
Recent methods explore Arm Shuffle in MA-VLA~\citep{zhang2026mavla} and arm-specific representations in TwinVLA~\citep{im2026twinvla} and SkillVLA~\citep{zhai2026skillvla}.
However, differences in backbones, datasets, and evaluation settings make their effects on compositional generalization difficult to isolate.
This motivates a controlled study: \emph{which training and architectural choices help familiar arm-level skills generalize to unseen compositions?}

We introduce \textbf{ACG-Bench}, comprising 23 task--condition pairs across 8 task families: 6 in-domain conditions and 17 unseen compositions covering \textbf{reordering}, \textbf{synchronization}, \textbf{Sync+Reorder}, and \textbf{cross-task composition}.
All methods receive identical per-arm atomic prompts from a common phase scheduler, isolating execution of a supplied composition from skill planning.
Success requires both the task goal and compliance with physical milestones and specified event order or timing, rather than merely reaching the final state.

We first compare three execution baselines.
All three use the same pretrained $\pi_{0.5}$ backbone~\citep{intelligence2025pi}, source data, and evaluation protocol.
The first is the standard \textbf{$\pi_{0.5}$}, where a single shared model jointly controls both arms.
The second is MA-VLA~\citep{zhang2026mavla}, which retains this shared-model setup.
It introduces Arm Shuffle during training to augment arm-role assignments.
The third is an independent-control baseline, which instead deploys a separate $\pi_{0.5}$ policy for each arm.
Each policy produces actions only for its assigned arm.
These comparisons examine whether arm-role augmentation or independent control improves compositional execution over the standard shared policy.
Despite their different designs, all three baselines show limited generalization to unseen compositions.

Starting from the standard shared-model $\pi_{0.5}$, we then investigate three targeted architectural changes: \textbf{arm-token grouping} for arm-specific representations, \textbf{skill-specific LoRA adapters (SkillLoRA)} for skill-conditioned parameters, and \textbf{arm-wise attention (AWA)} for regulating cross-arm information flow.
Individual changes yield modest gains, whereas their combination, \textbf{AE-VLA}, achieves 21.53\% generalization success in simulation and 39.00\% mean success across five unseen conditions on physical SO101 robots, compared with the strongest baseline means of 5.53\% and 10.00\%, respectively.
These findings suggest that modest architectural changes can substantially improve arm-wise compositional generalization within a shared policy, building on $\pi_{0.5}$'s pretrained capabilities.

Our contributions are threefold:
\begin{itemize}
    \item \textbf{ACG-Bench:} a benchmark for systematic evaluation of dual-arm skill recomposition across changes in order, synchronization, and cross-task combinations.
    \item \textbf{A unified empirical study:} comparisons of representative augmentation-based and architectural strategies using a common $\pi_{0.5}$ backbone, source dataset, and execution protocol.
    \item \textbf{Design insights and AE-VLA:} evidence of complementary benefits from skill-conditioned parameters and arm-wise attention, validated through their combined configuration in simulation and on physical robots.
\end{itemize}

\section{Related Work}
\label{sec:relatedwork}

\paragraph{Vision-language-action policies.}
VLA models map visual observations and language to robot actions, with recent
systems improving task, scene, and embodiment transfer through large robot
datasets, pretrained representations, and generative action
heads~\cite{brohan2022rt,zitkovich2023rt,team2024octo,kim2024openvla,black2024pi_0,pertsch2025fast,intelligence2025pi,wen2025dexvla}.
Several systems support bimanual control~\cite{nvidia2025gr00tn1openfoundation,liu2026rdt2},
and steerable policies study task composition through prompts or
subgoals~\cite{intelligence2026pi}. We focus on a different source of shift:
the local behaviors remain familiar, but the order or timing of skills across the two
arms changes. We study this shift on $\pi_{0.5}$.

\paragraph{Bimanual learning and coordination.}
Bimanual policies have been studied through joint action prediction,
cross-arm attention, and explicit coordination modules~\cite{grotz2024peract2,chernyadev2024bigym,lai2025roboballet,glossop2025cast}.
ACG asks whether the learned coordination can support relations absent
from training. We test this with shared scene context and restricted
attention between local arm streams.

\paragraph{Arm-wise representation and skill reuse.}
TwinVLA~\citep{im2026twinvla} represents shared and arm-specific inputs through
twin VLM streams with joint attention, sharing its visual encoder and action
head. SkillVLA~\citep{zhai2026skillvla} uses hierarchical skill decisions,
per-arm action generation, and adaptive communication for cooperative
behaviors. MA-VLA~\citep{zhang2026mavla} studies atomic instruction assignment
and arm-role permutation for compositional generalization.
We study the related ideas of per-arm token groups, skill-specific weights,
and attention access under one training and evaluation setup.
Appendix~\ref{appendix:design_correspondence} details the differences from
the complete prior systems.

\paragraph{Manipulation benchmarks and skill composition.}
Existing benchmarks evaluate generalization over objects, layouts, goals,
visual appearance, task sequences, and skill combinations~\cite{james2020rlbench,mees2022calvin,mu2021maniskill,liu2023libero,haresh2024clevrskills,zhang2025vlabench}.
Recent bimanual and multi-robot suites expand task and scene
diversity~\cite{mu2025robotwin,chen2025robotwin,wang2025roboeval,wu2025robocoin,qin2025robofactory,chen2026rmbench}.
ACG-Bench controls the low-level skill vocabulary and changes the arm-level
event graph itself. Its constraint-aware metric also rejects executions that
reach the final state through the wrong order or timing, making coordination
structure an explicit evaluation target.

\section{Arm-wise Compositional Generalization}
\label{sec:macg}

\subsection{Problem Formulation}
\label{sec:problem_formulation}

We study whether a dual-arm policy can execute familiar skills in combinations
absent from training. A bowl routine demonstrated left-arm-first may be
tested right-arm-first or with synchronized picks. Cross-task composition
combines behaviors from different source tasks, such as bowl handling and
cube placement.

We describe a task by an \emph{event graph}. Each node is a skill execution
with a specified arm, object, and target, such as ``left arm picks bowl A.''
Order constraints specify which skill events must happen before others.
Synchronization constraints require two designated completion milestones to
occur within a task-specific time window.

A test condition evaluates ACG when its required skill types occur in the
training demonstrations, but their joint event graph does not.
The robot setup stays fixed. The policy must achieve the task goal while
following the required order and timing. Arm-specific phase instructions
supply the target structure, so the question is whether the policy can
\emph{execute} a new composition of learned skills.

\subsection{ACG-Bench}

ACG-Bench builds dual-arm tasks in ManiSkill~\cite{mu2021maniskill} using the
task and expert-program generation pipeline of RoboTwin~2.0
~\cite{chen2025robotwin}. The benchmark has six source task families: Stack
Bowls, Stack Cubes, Push Cubes, Burger \& Fries, Can in Basket, and Object in
Cabinet. Cube in Bowl and Bowl in Cabinet are held-out cross-task families.
Task programs are run from varied valid initial states and repaired after
failed grasps, collisions, or failed task checks. Accepted programs collect
100 clean expert demonstrations per source task. We split them into skills
such as Pick, Place, Lift, Pull, Move, Push, Recover, and Wait. State checks
on grasp attachment, object location, and cabinet joints define skill
completion. Object and target arguments can change without creating a new
skill type. Appendix~\ref{appendix:atomic_skills} gives skill definitions,
source-task coverage, and example phase prompts.

We pool all 600 source demonstrations into one multitask training set.
Each method uses the same trained policy across all 23 task--condition pairs,
without task-specific fine-tuning; Dual $\pi_{0.5}$ uses a fixed pair of
per-arm policies. Held-out compositions contribute no training demonstrations.

At evaluation, a common rule-based scheduler supplies per-arm atomic prompts
to \emph{every} method. Each prompt gives the active skill and its arguments.
Shared state checks update prompts and record milestone completion times
for scoring. The vocabulary and transition rules are identical across methods.
This protocol tests execution of a supplied skill plan, not plan inference.

We construct four separate generalization groups:
\begin{itemize}
    \item \textbf{Reorder} reverses or changes event order
    while retaining the task goal and its skill events.
    \item \textbf{Sync} requires designated events of the two arms to complete
    within a time window instead of following the training sequence.
    \item \textbf{Sync+Reorder} changes order and timing rules together.
    \item \textbf{Cross-task Combination} combines skills from different
    source task families, as in Cube in Bowl and Bowl in Cabinet.
\end{itemize}
Each condition is counted in one reporting group. Cross-task conditions keep
that label even if they also change timing.
The resulting common set has 23 task--condition pairs: six in-domain, four
Reorder, six Sync, four Sync+Reorder, and three Cross-task Combination. The 17
non-in-domain pairs form the generalization set.

Cross-task tests also change the task context, while Sync+Reorder changes
both order and timing. The groups can therefore differ in more than one
aspect of difficulty. We report each group as well as the mean over all
17 generalization conditions. Table~\ref{tab:macg_task_results} lists every
condition in the main text.

\section{Design Choices on \texorpdfstring{$\pi_{0.5}$}{pi0.5}}
\label{sec:method}

\begin{figure*}[t]
    \centering
    \includegraphics[width=\linewidth]{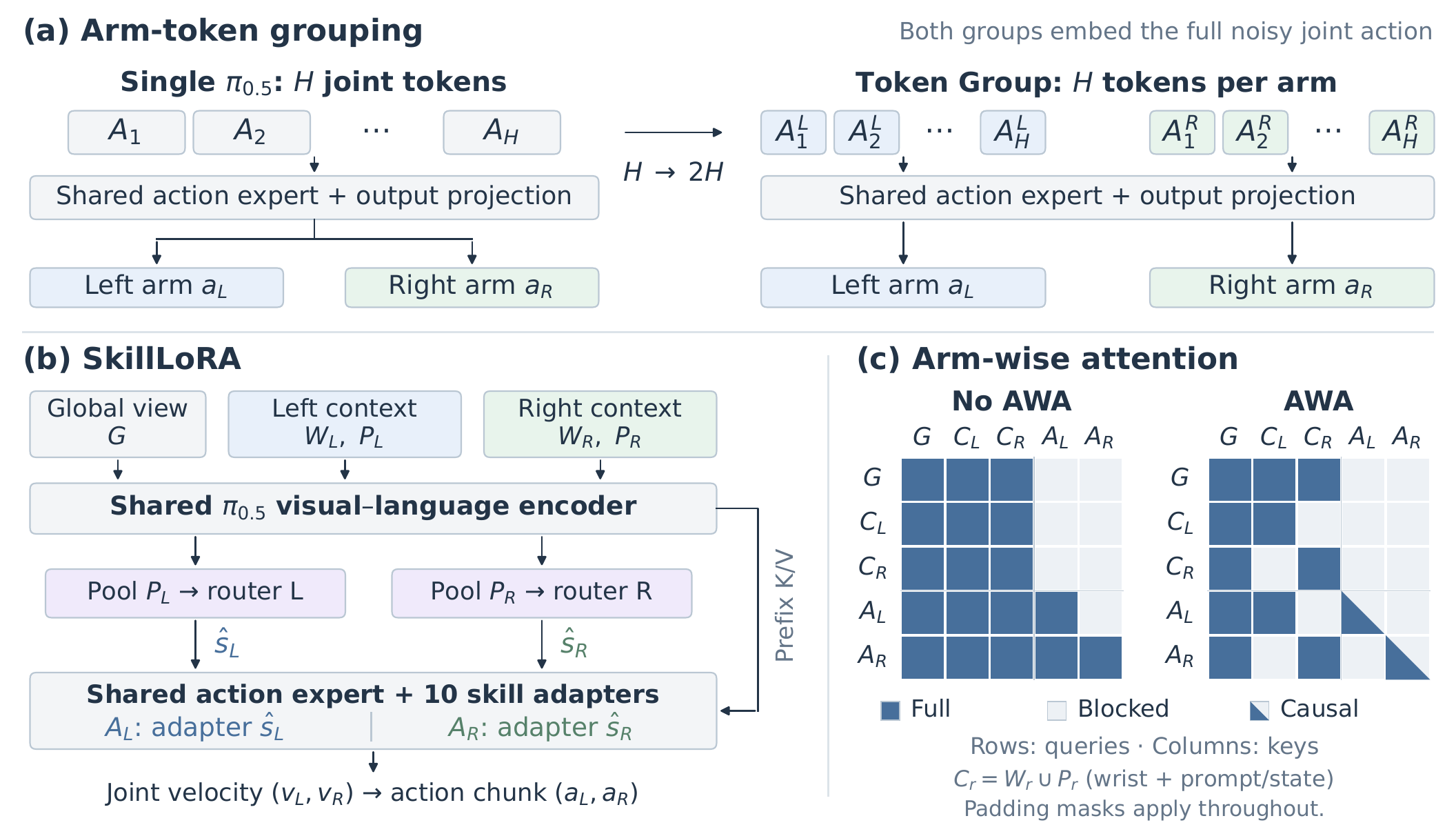}
    \caption{\textbf{Three design choices within one $\pi_{0.5}$ policy.}
    (a) Single uses $H$ joint action tokens to predict both arms. Token Group
    uses two groups of $H$ tokens, each assigned to one arm, within a shared
    action expert. (b) SkillLoRA selects a skill adapter for each group.
    (c) AWA controls attention between groups while retaining global context.
    Here $H=50$; arrows in (a) summarize decoding and denoising.}
    \label{fig:method}
\end{figure*}

We study token grouping, skill-specific weights, and attention on pretrained
$\pi_{0.5}$~\citep{intelligence2025pi_}.
Composing familiar skills under new coordination requirements calls for
reusing each arm's behavior while retaining information about the shared
scene. Arm-token grouping gives each arm an explicit action representation
within one shared backbone. SkillLoRA associates parameter updates with
atomic skills so they can be reused across arms and tasks. AWA restricts
direct attention to each arm's local inputs and actions while preserving
shared global context, aiming to reduce dependence on familiar arm
pairings without removing information needed for coordination.
We test all four SkillLoRA/AWA settings on Token Group and call the full
combination \textbf{AE-VLA}.

\subsection{Shared Backbone and Arm-token Grouping}

Every variant receives a global image $I_G$, wrist images $I_L,I_R$,
arm states $q_L,q_R$, and the same benchmark-provided active
instructions $p_L,p_R$. One policy generates a joint action chunk
$a=(a_L,a_R)$ over $H=50$ control steps. We retain the backbone's visual
encoder, vision--language model, and flow-matching action expert. At denoising
time $t$, the expert predicts a joint velocity field
\begin{equation}
    v_\theta(a_t,t\mid I_G,I_L,I_R,q_L,q_R,p_L,p_R)
    = (v_L,v_R),
    \label{eq:joint_policy}
\end{equation}
which is integrated to obtain both arms' actions.

Single $\pi_{0.5}$ uses $H$ joint action tokens: each token predicts both
arms at one chunk position. \textbf{Token Group} uses two groups of $H$
tokens, $[A_L,A_R]$, where $A_L$ supplies the left-arm output and $A_R$
the right-arm output (Figure~\ref{fig:method}a). Both groups pass through
one shared action expert and output projection. This gives each arm an
explicit token range on which SkillLoRA can select weights and AWA can
set attention connections.

The prefix is $[G,W_L,W_R,P_L,P_R]$: $G$ and $W_r$ are global and
wrist image tokens, and $P_r$ combines arm $r$'s prompt with its discrete
state. Fixed ranges and output assignments identify the arms, without a
learned arm-ID embedding. Both action groups still embed the full noisy
joint action, so grouping does not make their inputs independent.
Token Group alone uses the standard sequence mask; SkillLoRA and AWA
are added separately below. Appendix~\ref{appendix:attention_detail}
gives the exact layout and output assignment.

\subsection{Skill-conditioned Action Parameters}

\textbf{Token Group + SkillLoRA} adds a bank of $K=10$
low-rank updates indexed by atomic skill type. For a token $h$ in arm $r$'s
action range, an adapted linear transformation is
\begin{equation}
    h' = Wh + \frac{\alpha}{k} B_{\hat{s}_r}A_{\hat{s}_r}h,
    \qquad k=4,\quad \alpha=16,
    \label{eq:skill_adapter}
\end{equation}
where $W$ is shared and $\hat{s}_r$ is the selected skill. The banks are
shared across arms and tasks; the arms can select different skills or reuse
the same skill parameters.

Selection uses a separate two-layer router for each arm. The router pools
the prefix features at that arm's prompt/state positions after attention and
predicts a distribution over skill types; its argmax selects the adapter.
One selection is used for all $H$ tokens and denoising steps. Training and
inference both use the prediction; skill labels only supervise the routers.

SkillLoRA is applied to attention projections and feed-forward layers in
all 18 action-expert blocks, and to the action input/output projections.
The visual and language prefix has no skill adapters. This choice adds
both parameters and a router loss, so its effect cannot be assigned to
skill-specific weights alone.

\subsection{Arm-wise Attention Connectivity}

\textbf{Arm-wise attention (AWA)} applies to grouped policies with or
without SkillLoRA, giving \textbf{AE-VLA} and \textbf{Token Group + AWA}.
Let $C_r=W_r\cup P_r$ denote arm $r$'s local
prefix tokens, and $A_{r,j}$ its action token at chunk position $j$.
The allowed key sets $\mathcal{V}$ for each query group are
\begin{equation}
    \begin{aligned}
    \mathcal{V}(G) &= G\cup C_L\cup C_R,\\
    \mathcal{V}(C_r) &= G\cup C_r,\\
    \mathcal{V}(A_{r,j}) &= G\cup C_r\cup\{A_{r,i}:i\leq j\}.
    \end{aligned}
    \label{eq:arm_attention}
\end{equation}
The mask applies both within the observation prefix and from action queries
to prefix/action keys. Global queries read both local prefixes, so
cross-arm observation information can still pass through shared global
features. Prefix tokens do not attend to action tokens. No latent
collaboration tokens are used.

Without AWA, the standard sequence mask allows bidirectional attention
within each action block and lets a later block read earlier blocks.
AWA blocks both cross-arm action directions and makes each block lower
triangular (Figure~\ref{fig:method}c). Padding masks remain in both settings.
This comparison tests the whole mask, including its change to temporal
attention; Appendix~\ref{appendix:attention_detail} gives the details.

\subsection{Training and Comparison Protocol}

We retain flow matching, with $a_t=(1-t)a+t\epsilon$ and
$\epsilon\sim\mathcal{N}(0,I)$. SkillLoRA variants additionally optimize
the router's cross-entropy loss:
\begin{equation}
    \mathcal{L}
    = \mathbb{E}\!\left[\left\|v_\theta(a_t,t\mid o)-(\epsilon-a)\right\|^2\right]
    +0.1\,\mathcal{L}_{\mathrm{route}},
    \label{eq:method_loss}
\end{equation}
where $o$ denotes the common policy inputs and the action error is averaged
over chunk positions and action dimensions. The router loss averages
cross-entropy over valid per-arm skill labels; missing labels are ignored. The hard choice blocks action-loss gradients through the router, which
is trained by the classification loss. We jointly
fine-tune the shared backbone, action expert, adapters, and routers.
Variants without SkillLoRA use only the action loss.

\section{Experiments}
\label{sec:experiments}

\subsection{Experimental Setup}
\label{sec:experimental_setup}

\paragraph{Common pretrained backbone.}
Single $\pi_{0.5}$, both policies in Dual $\pi_{0.5}$, our MA-VLA implementation~\citep{zhang2026mavla}, and all four Token Group settings initialize their backbones from the same pretrained $\pi_{0.5}$ checkpoint~\citep{intelligence2025pi_}. This design keeps the pretraining source fixed across all comparisons. For MA-VLA, we replace the original $\pi_0$ backbone with $\pi_{0.5}$, and all reported MA-VLA results use this adapted implementation. In Dual $\pi_{0.5}$, each arm is controlled by an independent policy that receives its own wrist view together with the shared global view. We omit TwinVLA~\citep{im2026twinvla} from the quantitative comparison because its pretrained weights differ from those of $\pi_{0.5}$, which would confound architectural differences with differences in pretraining. These choices enable a more controlled comparison with respect to pretrained initialization.

\paragraph{Training and methods.}
Using the pooled source data in Section~\ref{sec:macg}, we train the
single-policy variants for 30,000 steps with batch size 32 on two NVIDIA
A800 GPUs.
Single $\pi_{0.5}$ predicts both arms with the base model.
Dual $\pi_{0.5}$ trains a separate policy for each arm.
All receive the same per-arm skill prompts at evaluation; SkillLoRA also
uses skill labels to train its routers.

\paragraph{Test set and averages.}
Each method runs 100 episodes per condition with seeds 1000--1099,
for 2,300 episodes across the 23 conditions. We average results equally
over conditions.
The generalization mean includes all 17 non-in-domain conditions, so the
four groups have weights $4{:}6{:}4{:}3$.

\paragraph{Metrics.}
Constraint-compliant success rate (CCSR) counts episodes that reach the
goal and meet the required milestones, order, and timing:
\begin{equation}
    \mathrm{CCSR}=100\times\frac{N_{\mathrm{success}}}{N_{\mathrm{episodes}}}.
    \label{eq:ccsr}
\end{equation}
Here, $N_{\mathrm{success}}$ denotes the number of valid successful episodes, while $N_{\mathrm{episodes}}$ denotes the total number of evaluated episodes. Normalized progress is defined as the fraction of required milestones completed under these scoring rules and is used to measure partial task execution. For Sync, the allowable event-time gap is 20\,s for Can in Basket, 15\,s for Object in Cabinet, and 45\,s for all other tasks. For Can in Basket, task completion is determined by environment success, while Sync additionally requires satisfaction of the 20\,s event-time constraint. If an arm-wise collision occurs, the corresponding episode is considered a failure. We therefore set its success score to zero while retaining the measured progress. Appendix~\ref{appendix:scoring} provides the complete scoring rules.

\subsection{Main Results}
\label{sec:main_results}

\begin{figure}[t]
    \centering
    \includegraphics[width=\linewidth]{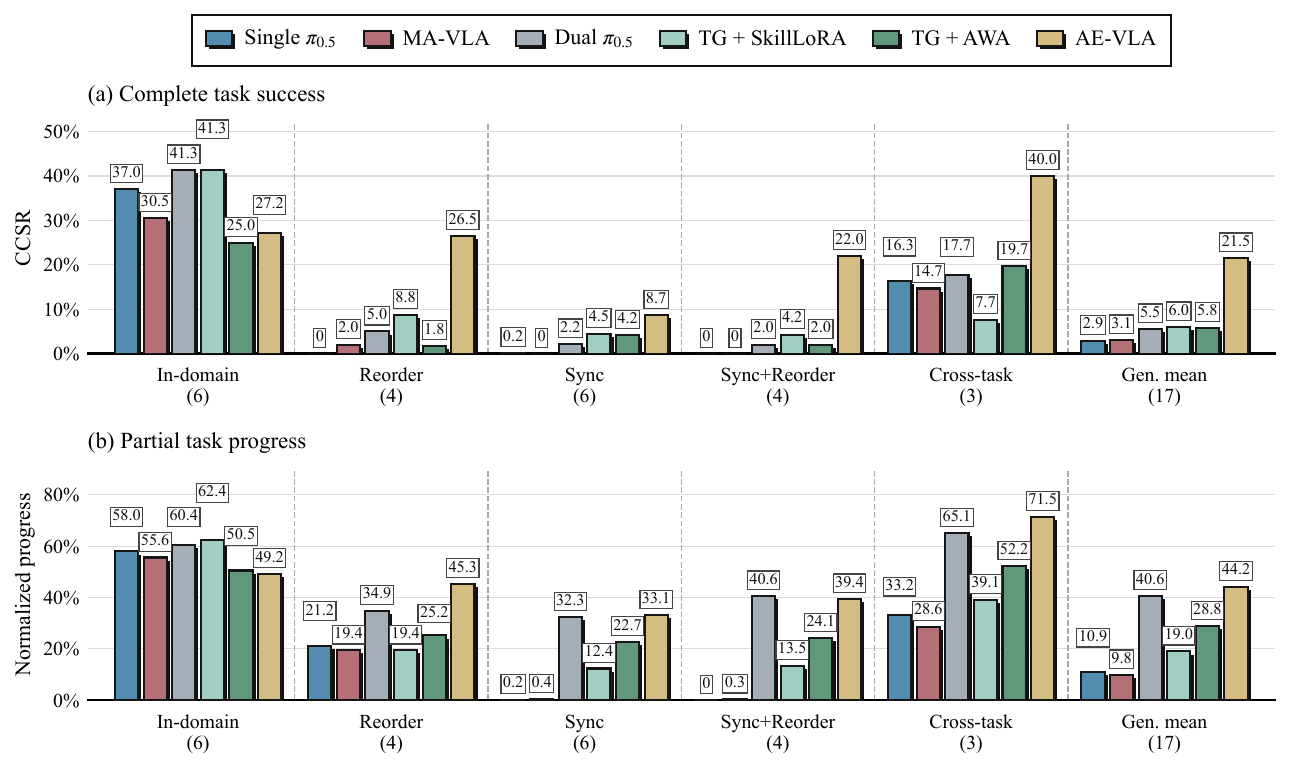}
    \caption{\textbf{Success and progress by test group.} Parentheses give
    condition counts; each condition has 100 episodes per method. TG means
    Token Group. Bars show means over conditions.
    Generalization averages all 17 non-in-domain conditions.}
    \label{fig:regime_results}
\end{figure}

Figure~\ref{fig:regime_results} shows a large gap between in-domain and
compositional success. Single $\pi_{0.5}$ falls from 37.00\% in-domain CCSR
to 2.94\% on the generalization set. MA-VLA reaches 3.06\% and Dual
$\pi_{0.5}$ reaches 5.53\% on this set. The combined AE-VLA setting reaches
21.53\%, a gain of 18.59 points over Single and 16.00 over Dual.
It reaches 26.50\% on Reorder, 22.00\% on Sync+Reorder, and 40.00\%
on Cross-task Combination. Pure Sync remains difficult: AE-VLA reaches
8.67\%. Its in-domain success is 27.17\%, below all three baselines.

Progress gives a different view. AE-VLA and Dual $\pi_{0.5}$ complete similar fractions of required milestones on the generalization set (44.25\% and 40.64\%), yet their full success rates differ widely. Thus, local progress does not imply successful completion of the required combination. For example, on synchronization tasks, Dual $\pi_{0.5}$ can grasp the objects but lacks the waiting and coordination needed between arms, causing collisions between arms and task failure. Single and MA-VLA also show low progress (10.94\% and 9.83\%).

Gains also vary across tasks. AE-VLA reaches 79\% on Stack Bowls/Reorder
versus 4\% for Dual, and 54\% on Cube in Bowl/Cross-task versus 0\%.
Dual is stronger on several other conditions, including Object in
Cabinet/In-domain and Burger \& Fries/Sync. Single is best on Push
Cubes/In-domain; MA-VLA is best on Stack Cubes/In-domain.
Table~\ref{tab:macg_task_results} reports all 23 conditions, including
cases where the baselines do better.

\begin{table*}[t]
\centering
\small
\setlength{\tabcolsep}{4pt}
\caption{\textbf{All simulation conditions: CCSR (\%).} Each condition has 100 episodes per method. Only Dual $\pi_{0.5}$ uses two policies. Parentheses in cross-task rows retain the timing setting. Bold marks the best result.}
\label{tab:macg_task_results}
\begin{tabular}{llrrrr}
\toprule
Task & Condition & \shortstack{Single\\$\pi_{0.5}$} & MA-VLA & \shortstack{Dual\\$\pi_{0.5}$} & AE-VLA \\
\midrule
Stack Bowls & In-domain & 74.00 & 77.00 & \textbf{83.00} & \textbf{83.00} \\
 & Reorder & 0.00 & 7.00 & 4.00 & \textbf{79.00} \\
 & Sync & 0.00 & 0.00 & 0.00 & \textbf{26.00} \\
 & Sync+Reorder & 0.00 & 0.00 & 1.00 & \textbf{67.00} \\
\midrule
Stack Cubes & In-domain & 17.00 & \textbf{28.00} & 27.00 & 20.00 \\
 & Reorder & 0.00 & 1.00 & 9.00 & \textbf{19.00} \\
 & Sync & 0.00 & 0.00 & \textbf{4.00} & 3.00 \\
 & Sync+Reorder & 0.00 & 0.00 & 0.00 & \textbf{21.00} \\
\midrule
Push Cubes & In-domain & \textbf{50.00} & 10.00 & 12.00 & 8.00 \\
 & Reorder & 0.00 & 0.00 & \textbf{6.00} & 3.00 \\
 & Sync & 0.00 & 0.00 & 2.00 & \textbf{4.00} \\
 & Sync+Reorder & 0.00 & 0.00 & \textbf{1.00} & 0.00 \\
\midrule
Burger \& Fries & In-domain & 9.00 & 0.00 & \textbf{17.00} & 6.00 \\
 & Reorder & 0.00 & 0.00 & 1.00 & \textbf{5.00} \\
 & Sync & 0.00 & 0.00 & \textbf{6.00} & 1.00 \\
 & Sync+Reorder & 0.00 & 0.00 & \textbf{6.00} & 0.00 \\
\midrule
Can in Basket & In-domain & \textbf{58.00} & 29.00 & 50.00 & 25.00 \\
 & Sync & 1.00 & 0.00 & 0.00 & \textbf{18.00} \\
\midrule
Cube in Bowl & Cross-task (Reference) & 0.00 & 3.00 & 0.00 & \textbf{54.00} \\
 & Cross-task (Sync) & 0.00 & 0.00 & 0.00 & \textbf{11.00} \\
\midrule
Object in Cabinet & In-domain & 14.00 & 39.00 & \textbf{59.00} & 21.00 \\
 & Sync & 0.00 & 0.00 & \textbf{1.00} & 0.00 \\
\midrule
Bowl in Cabinet & Cross-task (Reference) & 49.00 & 41.00 & 53.00 & \textbf{55.00} \\
\bottomrule
\end{tabular}
\end{table*}

\subsection{Design Choices and Their Interaction}
\label{sec:ablation}

\begin{table}[t]
\centering
\small
\setlength{\tabcolsep}{2.5pt}
\caption{\textbf{Design choices on the same $\pi_{0.5}$ backbone.} The four grouped variants form a $2\times2$ comparison of SkillLoRA and AWA. Values are percentages averaged over conditions.}
\label{tab:macg_ablation}
\begin{tabular}{lcccrrr}
\toprule
Variant & \shortstack{Token\\Group} & SkillLoRA & \shortstack{Arm-wise\\Attn} & \shortstack{ID\\CCSR} & \shortstack{Gen.\\CCSR} & \shortstack{Gen.\\Progress} \\
\midrule
Single $\pi_{0.5}$ & -- & -- & -- & 37.00 & 2.94 & 10.94 \\
Token Group & \checkmark & -- & -- & \textbf{45.00} & 3.82 & 10.98 \\
Token Group + SkillLoRA & \checkmark & \checkmark & -- & 41.33 & 6.00 & 19.02 \\
Token Group + AWA & \checkmark & -- & \checkmark & 25.00 & 5.82 & 28.84 \\
Token Group + Both (AE-VLA) & \checkmark & \checkmark & \checkmark & 27.17 & \textbf{21.53} & \textbf{44.25} \\
\bottomrule
\end{tabular}
\end{table}

Table~\ref{tab:macg_ablation} shows the complete comparison. Token Group
raises generalization CCSR from 2.94\% to 3.82\%, with almost no change
in progress. Adding SkillLoRA alone reaches 6.00\%; adding AWA alone
reaches 5.82\%. Their combination reaches 21.53\%. AWA alone has higher
progress than SkillLoRA alone (28.84\% versus 19.02\%), but not higher
full success.

The gain from AWA is 2.00 points without SkillLoRA and 15.53 with it.
The difference, 13.53 points, measures their interaction. The two choices work better
together for these trained models. Appendix~\ref{appendix:architecture_results}
gives the results for each condition.

This improvement has an in-domain cost. AWA lowers Token Group from
45.00\% to 25.00\% in-domain CCSR, and Token Group + SkillLoRA from
41.33\% to 27.17\%. The comparison also has limits: SkillLoRA adds both
weights and router supervision; AWA changes both cross-arm and within-chunk
attention. The four settings separate the effects of these two design
choices, but do not isolate each change within them.

\subsection{Real-world SO101 Results}
\label{sec:real_world}

\begin{figure*}[t]
    \centering
    \includegraphics[width=\linewidth]{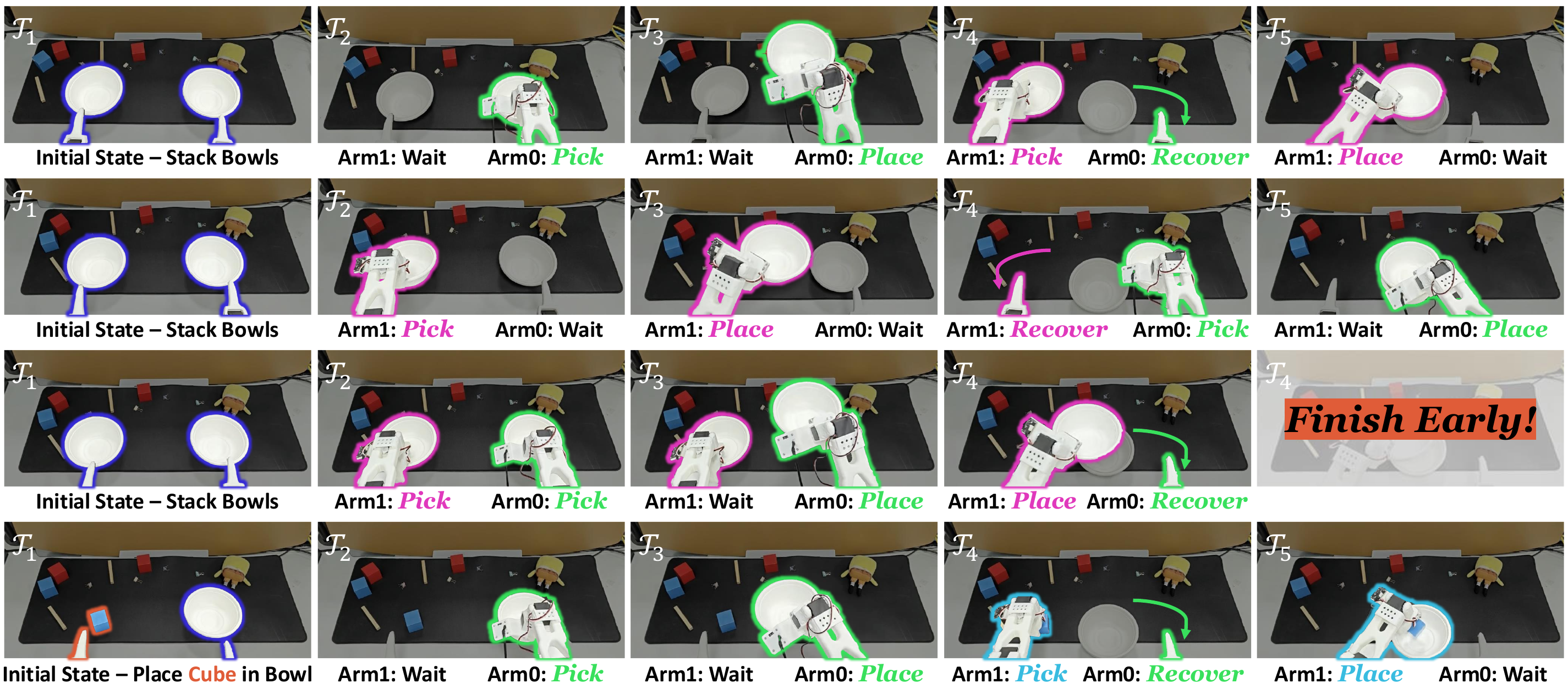}
    \caption{\textbf{SO101 examples.} Rows show Stack Bowls/In-domain,
    Stack Bowls/Reorder, Stack Bowls/Sync, and Cube in Bowl/Cross-task.
    Table~\ref{tab:real_robot_results} reports all evaluation trials.}
    \label{fig:real_robot}
\end{figure*}

We use two 6-DoF SO101 arms with one global and two wrist cameras.
Training demonstrations cover Stack Bowls, Stack Cubes, and Push Cubes,
collected with LeRobot teleoperation~\cite{cadene2024lerobot}.
Cube in Bowl is held out: its cube pick-and-place skill is learned from
Stack Cubes and tested with a bowl target. We evaluate three in-domain
and five unseen conditions with distractor objects. Each method has 20
trials per condition, giving 640 trials in total.
Appendix~\ref{appendix:so101} describes the setup and scoring.

\begin{table}[t]
\centering
\small
\setlength{\tabcolsep}{2.7pt}
\renewcommand{\arraystretch}{1.02}
\caption{\textbf{SO101 results with distractor objects.} Each task row
shows successes out of 20 trials. Mean rows average rates over three
in-domain or five unseen conditions. Bold marks the best result.}
\label{tab:real_robot_results}
\begin{tabular*}{\columnwidth}{@{\extracolsep{\fill}}llcccc}
\toprule
Setting & Task & \shortstack{Single\\$\pi_{0.5}$} & MA-VLA
& \shortstack{Dual\\$\pi_{0.5}$} & AE-VLA \\
\midrule
\multirow{3}{*}{In-domain}
& Stack Bowls & 15/20 & 15/20 & \textbf{18/20} & 17/20 \\
& Stack Cubes & 5/20 & 3/20 & \textbf{10/20} & 9/20 \\
& Push Cubes  & 10/20 & 9/20 & \textbf{14/20} & 10/20 \\
\cmidrule(lr){2-6}
& \textbf{ID mean (\%)} & 50.00 & 45.00 & \textbf{70.00} & 60.00 \\
\midrule
Reorder & Stack Bowls & 0/20 & 4/20 & 0/20 & \textbf{9/20} \\
Reorder & Push Cubes & 0/20 & 2/20 & 0/20 & \textbf{6/20} \\
Sync & Stack Bowls & 0/20 & 0/20 & 0/20 & \textbf{12/20} \\
Sync & Push Cubes & 0/20 & 0/20 & \textbf{6/20} & 2/20 \\
Cross-task & Cube in Bowl & 2/20 & 0/20 & 4/20 & \textbf{10/20} \\
\cmidrule(lr){2-6}
& \textbf{OOD mean (\%)} & 2.00 & 6.00 & 10.00 & \textbf{39.00} \\
\bottomrule
\end{tabular*}
\end{table}

AE-VLA reaches 39.00\% mean success on the five unseen conditions,
compared with 2.00\% for Single, 6.00\% for MA-VLA, and 10.00\% for Dual
(Table~\ref{tab:real_robot_results}). It leads on four of the five conditions,
including Stack Bowls/Sync (12/20 versus 0/20 for every baseline).
Push Cubes/Sync is the exception: Dual reaches 6/20 and AE-VLA reaches
2/20. In-domain, AE-VLA reaches 60.00\%, below Dual's 70.00\%.
The real-robot results thus also show improved generalization with some
loss of in-domain success. Figure~\ref{fig:real_robot} shows example phases.

\section{Conclusion}
\label{sec:conclusion}

We introduced \textbf{ACG-Bench}, a testbed for evaluating dual-arm skill reuse under unseen ordering, synchronization, and cross-task compositions. Using a common $\pi_{0.5}$ backbone, we compare representative data-augmentation and architectural strategies, highlighting the complementarity of skill-conditioned parameters and arm-wise attention. Their combination, \textbf{AE-VLA}, achieves 21.53\% generalization CCSR in simulation and 39.00\% mean success across five unseen SO101 conditions, versus 5.53\% and 10.00\% for Dual $\pi_{0.5}$.
Our study focuses on execution under supplied skill structures, while synchronization and in-domain performance remain challenging. A key next step is a high-level planner that decomposes novel tasks into familiar skills, assigns them to arms, determines order and synchronization, and revises plans from execution feedback. Coupling planning with compositional execution would enable evaluation of the full planning--execution loop and test whether systems can construct and execute new skill combinations for unseen tasks.
\label{sec:main_text_end}

\clearpage
\typeout{EML-MAIN-TEXT-PAGES: \the\numexpr\value{page}-1\relax}
\section*{AI Use Statement}
Generative AI tools were used solely to polish the language and presentation of this manuscript,
including grammar, phrasing, and readability improvements. They were not used to generate ex-
perimental data, design the benchmark or evaluation protocol, implement the evaluation methods,
or replace the authors’ scientific judgment. The authors reviewed all AI-assisted text and take full
responsibility for the accuracy, claims, results, and final content of this pape

\section*{Ethics Statement}
This work studies compositional generalization in simulated and physical
dual-arm manipulation. We do not anticipate specific ethical concerns from
the reported research beyond the standard safety risks of physical robot
operation, which require appropriate precautions in deployment.

\section*{Reproducibility Statement}
Appendices~\ref{appendix:details}--\ref{appendix:evaluation} describe the
benchmark conditions and atomic skills, policy implementation and training
configuration, and evaluation protocols. They also provide per-condition
architectural results and the real-world experimental setup to support
reproduction and comparison.ility for the final content.

\bibliography{references}

@article{chen2025robotwin,
  title={Robotwin 2.0: A scalable data generator and benchmark with strong domain randomization for robust bimanual robotic manipulation},
  author={Chen, Tianxing and Chen, Zanxin and Chen, Baijun and Cai, Zijian and Liu, Yibin and Li, Zixuan and Liang, Qiwei and Lin, Xianliang and Ge, Yiheng and Gu, Zhenyu and others},
  journal={arXiv preprint arXiv:2506.18088},
  year={2025}
}

@article{black2024pi_0,
  title={pi0: A Vision-Language-Action Flow Model for General Robot Control},
  author={Black, Kevin and Brown, Noah and Driess, Danny and Esmail, Adnan and Equi, Michael and Finn, Chelsea and Fusai, Niccolo and Groom, Lachy and Hausman, Karol and Ichter, Brian and others},
  journal={arXiv preprint arXiv:2410.24164},
  year={2024}
}

@article{lai2025roboballet,
  title={RoboBallet: Planning for multirobot reaching with graph neural networks and reinforcement learning},
  author={Lai, Matthew and Go, Keegan and Li, Zhibin and Kr{\"o}ger, Torsten and Schaal, Stefan and Allen, Kelsey and Scholz, Jonathan},
  journal={Science Robotics},
  volume={10},
  number={106},
  pages={eads1204},
  year={2025},
  publisher={American Association for the Advancement of Science}
}

@article{intelligence2025pi_,
  title={pi05: a Vision-Language-Action Model with Open-World Generalization},
  author={Intelligence, Physical and Black, Kevin and Brown, Noah and Darpinian, James and Dhabalia, Karan and Driess, Danny and Esmail, Adnan and Equi, Michael and Finn, Chelsea and Fusai, Niccolo and others},
  journal={arXiv preprint arXiv:2504.16054},
  year={2025}
}

@article{intelligence2025pi,
  title={pi06: a VLA That Learns From Experience},
  author={Intelligence, Physical and Amin, Ali and Aniceto, Raichelle and Balakrishna, Ashwin and Black, Kevin and Conley, Ken and Connors, Grace and Darpinian, James and Dhabalia, Karan and DiCarlo, Jared and others},
  journal={arXiv preprint arXiv:2511.14759},
  year={2025}
}

@article{pertsch2025fast,
  title={Fast: Efficient action tokenization for vision-language-action models},
  author={Pertsch, Karl and Stachowicz, Kyle and Ichter, Brian and Driess, Danny and Nair, Suraj and Vuong, Quan and Mees, Oier and Finn, Chelsea and Levine, Sergey},
  journal={arXiv preprint arXiv:2501.09747},
  year={2025}
}

@article{james2020rlbench,
  title={Rlbench: The robot learning benchmark \& learning environment},
  author={James, Stephen and Ma, Zicong and Arrojo, David Rovick and Davison, Andrew J},
  journal={IEEE Robotics and Automation Letters},
  volume={5},
  number={2},
  pages={3019--3026},
  year={2020},
  publisher={IEEE}
}

@article{mees2022calvin,
  title={Calvin: A benchmark for language-conditioned policy learning for long-horizon robot manipulation tasks},
  author={Mees, Oier and Hermann, Lukas and Rosete-Beas, Erick and Burgard, Wolfram},
  journal={IEEE Robotics and Automation Letters},
  volume={7},
  number={3},
  pages={7327--7334},
  year={2022},
  publisher={IEEE}
}

@article{mu2021maniskill,
  title={Maniskill: Generalizable manipulation skill benchmark with large-scale demonstrations},
  author={Mu, Tongzhou and Ling, Zhan and Xiang, Fanbo and Yang, Derek and Li, Xuanlin and Tao, Stone and Huang, Zhiao and Jia, Zhiwei and Su, Hao},
  journal={arXiv preprint arXiv:2107.14483},
  year={2021}
}

@article{liu2023libero,
  title={Libero: Benchmarking knowledge transfer for lifelong robot learning},
  author={Liu, Bo and Zhu, Yifeng and Gao, Chongkai and Feng, Yihao and Liu, Qiang and Zhu, Yuke and Stone, Peter},
  journal={Advances in Neural Information Processing Systems},
  volume={36},
  pages={44776--44791},
  year={2023}
}

@inproceedings{zhang2025vlabench,
  title={Vlabench: A large-scale benchmark for language-conditioned robotics manipulation with long-horizon reasoning tasks},
  author={Zhang, Shiduo and Xu, Zhe and Liu, Peiju and Yu, Xiaopeng and Li, Yuan and Gao, Qinghui and Fei, Zhaoye and Yin, Zhangyue and Wu, Zuxuan and Jiang, Yu-Gang and others},
  booktitle={Proceedings of the IEEE/CVF International Conference on Computer Vision},
  pages={11142--11152},
  year={2025}
}

@article{haresh2024clevrskills,
  title={Clevrskills: Compositional language and visual reasoning in robotics},
  author={Haresh, Sanjay and Dijkman, Daniel and Bhattacharyya, Apratim and Memisevic, Roland},
  journal={Advances in Neural Information Processing Systems},
  volume={37},
  pages={38235--38266},
  year={2024}
}

@inproceedings{mu2025robotwin,
  title={Robotwin: Dual-arm robot benchmark with generative digital twins},
  author={Mu, Yao and Chen, Tianxing and Chen, Zanxin and Peng, Shijia and Lan, Zhiqian and Gao, Zeyu and Liang, Zhixuan and Yu, Qiaojun and Zou, Yude and Xu, Mingkun and others},
  booktitle={Proceedings of the computer vision and pattern recognition conference},
  pages={27649--27660},
  year={2025}
}

@article{wang2025roboeval,
  title={Roboeval: Where robotic manipulation meets structured and scalable evaluation},
  author={Wang, Yi Ru and Ung, Carter and Tannert, Grant and Duan, Jiafei and Li, Josephine and Le, Amy and Oswal, Rishabh and Grotz, Markus and Pumacay, Wilbert and Deng, Yuquan and others},
  journal={arXiv preprint arXiv:2507.00435},
  year={2025}
}

@inproceedings{qin2025robofactory,
  title={Robofactory: Exploring embodied agent collaboration with compositional constraints},
  author={Qin, Yiran and Kang, Li and Song, Xiufeng and Yin, Zhenfei and Liu, Xiaohong and Liu, Xihui and Zhang, Ruimao and Bai, Lei},
  booktitle={Proceedings of the IEEE/CVF International Conference on Computer Vision},
  pages={10075--10085},
  year={2025}
}

@article{kim2024openvla,
  title={Openvla: An open-source vision-language-action model},
  author={Kim, Moo Jin and Pertsch, Karl and Karamcheti, Siddharth and Xiao, Ted and Balakrishna, Ashwin and Nair, Suraj and Rafailov, Rafael and Foster, Ethan and Lam, Grace and Sanketi, Pannag and others},
  journal={arXiv preprint arXiv:2406.09246},
  year={2024}
}

@article{intelligence2026pi,
  title={pi07: a Steerable Generalist Robotic Foundation Model with Emergent Capabilities},
  author={Intelligence, Physical and Ai, Bo and Amin, Ali and Aniceto, Raichelle and Balakrishna, Ashwin and Balke, Greg and Black, Kevin and Bokinsky, George and Cao, Shihao and Charbonnier, Thomas and others},
  journal={arXiv preprint arXiv:2604.15483},
  year={2026}
}

@article{glossop2025cast,
  title={Cast: Counterfactual labels improve instruction following in vision-language-action models},
  author={Glossop, Catherine and Chen, William and Bhorkar, Arjun and Shah, Dhruv and Levine, Sergey},
  journal={arXiv preprint arXiv:2508.13446},
  year={2025}
}

@article{brohan2022rt,
  title={Rt-1: Robotics transformer for real-world control at scale},
  author={Brohan, Anthony and Brown, Noah and Carbajal, Justice and Chebotar, Yevgen and Dabis, Joseph and Finn, Chelsea and Gopalakrishnan, Keerthana and Hausman, Karol and Herzog, Alex and Hsu, Jasmine and others},
  journal={arXiv preprint arXiv:2212.06817},
  year={2022}
}

@inproceedings{zitkovich2023rt,
  title={Rt-2: Vision-language-action models transfer web knowledge to robotic control},
  author={Zitkovich, Brianna and Yu, Tianhe and Xu, Sichun and Xu, Peng and Xiao, Ted and Xia, Fei and Wu, Jialin and Wohlhart, Paul and Welker, Stefan and Wahid, Ayzaan and others},
  booktitle={Conference on Robot Learning},
  pages={2165--2183},
  year={2023},
  organization={PMLR}
}

@article{team2024octo,
  title={Octo: An open-source generalist robot policy},
  author={Team, Octo Model and Ghosh, Dibya and Walke, Homer and Pertsch, Karl and Black, Kevin and Mees, Oier and Dasari, Sudeep and Hejna, Joey and Kreiman, Tobias and Xu, Charles and others},
  journal={arXiv preprint arXiv:2405.12213},
  year={2024}
}

@article{liu2026rdt2,
  title={RDT2: Exploring the Scaling Limit of UMI Data Towards Zero-Shot Cross-Embodiment Generalization},
  author={Liu, Songming and Li, Bangguo and Ma, Kai and Wu, Lingxuan and Tan, Hengkai and Ouyang, Xiao and Su, Hang and Zhu, Jun},
  journal={arXiv preprint arXiv:2602.03310},
  year={2026}
}

@misc{nvidia2025gr00tn1openfoundation,
      title={GR00T N1: An Open Foundation Model for Generalist Humanoid Robots}, 
      author={NVIDIA and Johan Bjorck and Fernando Castañeda and Nikita Cherniadev and Xingye Da and Runyu Ding and Linxi "Jim" Fan and Yu Fang and Dieter Fox and Fengyuan Hu and Spencer Huang and Joel Jang and Zhenyu Jiang and Jan Kautz and Kaushil Kundalia and Lawrence Lao and Zhiqi Li and Zongyu Lin and Kevin Lin and Guilin Liu and Edith Llontop and Loic Magne and Ajay Mandlekar and Avnish Narayan and Soroush Nasiriany and Scott Reed and You Liang Tan and Guanzhi Wang and Zu Wang and Jing Wang and Qi Wang and Jiannan Xiang and Yuqi Xie and Yinzhen Xu and Zhenjia Xu and Seonghyeon Ye and Zhiding Yu and Ao Zhang and Hao Zhang and Yizhou Zhao and Ruijie Zheng and Yuke Zhu},
      year={2025},
      eprint={2503.14734},
      archivePrefix={arXiv},
      primaryClass={cs.RO},
}

@article{wen2025dexvla,
  title={Dexvla: Vision-language model with plug-in diffusion expert for general robot control},
  author={Wen, Junjie and Zhu, Yichen and Li, Jinming and Tang, Zhibin and Shen, Chaomin and Feng, Feifei},
  journal={arXiv preprint arXiv:2502.05855},
  year={2025}
}

@article{chen2026rmbench,
  title={RMBench: Memory-Dependent Robotic Manipulation Benchmark with Insights into Policy Design},
  author={Chen, Tianxing and Wang, Yuran and Li, Mingleyang and Qin, Yan and Shi, Hao and Li, Zixuan and Hu, Yifan and Zhang, Yingsheng and Wang, Kaixuan and Chen, Yue and others},
  journal={arXiv preprint arXiv:2603.01229},
  year={2026}
}

@inproceedings{grotz2024peract2,
  title={Peract2: Benchmarking and learning for robotic bimanual manipulation tasks},
  author={Grotz, Markus and Shridhar, Mohit and Chao, Yu-Wei and Asfour, Tamim and Fox, Dieter},
  booktitle={CoRL 2024 Workshop on Whole-body Control and Bimanual Manipulation: Applications in Humanoids and Beyond},
  year={2024}
}

@article{chernyadev2024bigym,
  title={Bigym: A demo-driven mobile bi-manual manipulation benchmark},
  author={Chernyadev, Nikita and Backshall, Nicholas and Ma, Xiao and Lu, Yunfan and Seo, Younggyo and James, Stephen},
  journal={arXiv preprint arXiv:2407.07788},
  year={2024}
}

@article{wu2025robocoin,
  title={RoboCOIN: An Open-Sourced Bimanual Robotic Data COllection for INtegrated Manipulation},
  author={Wu, Shihan and Liu, Xuecheng and Xie, Shaoxuan and Wang, Pengwei and Li, Xinghang and Yang, Bowen and Li, Zhe and Zhu, Kai and Wu, Hongyu and Liu, Yiheng and others},
  journal={arXiv preprint arXiv:2511.17441},
  year={2025}
}

@misc{cadene2024lerobot,
    author = {Cadene, Remi and Alibert, Simon and Soare, Alexander and Gallouedec, Quentin and Zouitine, Adil and Palma, Steven and Kooijmans, Pepijn and Aractingi, Michel and Shukor, Mustafa and Aubakirova, Dana and Russi, Martino and Capuano, Francesco and Pascal, Caroline and Choghari, Jade and Moss, Jess and Wolf, Thomas},
    title = {LeRobot: State-of-the-art Machine Learning for Real-World Robotics in Pytorch},
    howpublished = "\url{https://github.com/huggingface/lerobot}",
    year = {2024}
}

@inproceedings{im2026twinvla,
  title={{TwinVLA}: Data-Efficient Bimanual Manipulation with Twin Single-Arm Vision-Language-Action Models},
  author={Im, Hokyun and Jeong, Euijin and Kolobov, Andrey and Fu, Jianlong and Lee, Youngwoon},
  booktitle={International Conference on Learning Representations},
  year={2026},
  url={https://arxiv.org/abs/2511.05275}
}

@article{zhai2026skillvla,
  title={{SkillVLA}: Tackling Combinatorial Diversity in Dual-Arm Manipulation via Skill Reuse},
  author={Zhai, Xuanran and Huang, Zekai and Wu, Longyan and Zhao, Qianyou and Yu, Qiaojun and Ren, Jieji and Hao, Ce and Soh, Harold},
  journal={arXiv preprint arXiv:2603.03836},
  year={2026},
  url={https://arxiv.org/abs/2603.03836}
}

@article{wu2026atombench,
  title={{ATOM-Bench}: A Real-World Benchmark for Atomic Skills and Compositional Generalization in Manipulation Policies},
  author={Wu, Zenan and Wei, Bingqing and Liu, Lu and He, Zheqi and Wang, Xi and Liu, Jiakang and Li, Zehui and Yao, Guocai and Zheng, Jing-Shu and Yang, Xi and Wang, Yongtao},
  journal={arXiv preprint arXiv:2606.16826},
  year={2026},
  url={https://arxiv.org/abs/2606.16826}
}

@inproceedings{jiang2023vima,
  title={{VIMA}: General Robot Manipulation with Multimodal Prompts},
  author={Jiang, Yunfan and Gupta, Agrim and Zhang, Zichen and Wang, Guanzhi and Dou, Yongqiang and Chen, Yanjun and Fei-Fei, Li and Anandkumar, Anima and Zhu, Yuke and Fan, Linxi},
  booktitle={International Conference on Machine Learning},
  year={2023},
  url={https://arxiv.org/abs/2210.03094}
}

@article{zhang2026mavla,
  title={{MA-VLA}: Multi-Arm Vision-Language-Action Model for Collaboration and Compositional Generalization},
  author={Zhang, Zaibin and Xiao, Junlan and Zhang, Zhongbo and Wang, Yifan and Kang, Li and Qin, Yiran and Xia, Changxing and Zhou, Heng and Fu, Talas and Zhou, Enshen and Zhang, Ruimao and Yin, Zhenfei and Lu, Huchuan and Wang, Lijun},
  journal={arXiv preprint arXiv:2608.25864},
  year={2026},
  url={https://arxiv.org/abs/2608.25864}
}

@article{chen2026robodojo,
  title={RoboDojo: A unified sim-and-real benchmark for comprehensive evaluation of generalist robot manipulation policies},
  author={Chen, Tianxing and Chen, Yue and Li, Zixuan and Tang, Junyuan and Su, Kailun and Lu, Haoran and Wan, Weijie and Chen, Baijun and Liu, Songling and Yan, Haowen and others},
  journal={arXiv preprint arXiv:2607.04434},
  year={2026}
}
\bibliographystyle{plainnat}

\clearpage
\appendix
\section{Benchmark and Atomic Skills}
\label{appendix:details}

We first describe the conditions, atomic operations, and phase prompts.
We then give the policy implementation, complete results, and scoring rules.

\subsection{Condition Inventory}

Table~\ref{tab:condition_inventory} lists the 23 shared conditions.
Each cross-task condition is counted in the Cross-task Combination group,
even when it also changes timing.

\begin{table}[htbp]
\centering
\small
\caption{ACG-Bench condition inventory. Reference and Sync distinguish
the execution requirements within the Cross-task group.}
\label{tab:condition_inventory}
\begin{tabular}{lccccc}
\toprule
Task family & ID & Reorder & Sync & Sync+R & Cross-task \\
\midrule
Stack Bowls & \checkmark & \checkmark & \checkmark & \checkmark & -- \\
Stack Cubes & \checkmark & \checkmark & \checkmark & \checkmark & -- \\
Push Cubes & \checkmark & \checkmark & \checkmark & \checkmark & -- \\
Burger \& Fries & \checkmark & \checkmark & \checkmark & \checkmark & -- \\
Can in Basket & \checkmark & -- & \checkmark & -- & -- \\
Cube in Bowl & -- & -- & -- & -- & Reference, Sync \\
Object in Cabinet & \checkmark & -- & \checkmark & -- & -- \\
Bowl in Cabinet & -- & -- & -- & -- & Reference \\
\midrule
Number of conditions & 6 & 4 & 6 & 4 & 3 \\
\bottomrule
\end{tabular}
\end{table}

\subsection{Task-program and Demonstration Generation}

Each task starts with a goal, scene, object list, allowed motion-planning
calls, event graph, and success checks. A code-generation agent writes the
simulator task and expert program. We run the program from varied valid
initial states. A checking agent identifies collisions, failed grasps, and
failed task checks, and the program is revised. Accepted programs collect
100 clean expert demonstrations per source task. We split the trajectories
into skills using gripper state, object attachment, object and end-effector
poses, and joint positions.

\subsection{Atomic Skills and Their Arguments}
\label{appendix:atomic_skills}

An atomic skill describes what one arm should do during a phase. It is
smaller than a complete dual-arm task, but can include several control steps:
Pick includes approaching and grasping an object; Place includes carrying
it to a target and releasing it. We describe a skill execution as
$(r,s,o,g)$, where $r$ is the arm, $s$ the operation, $o$ the object or
articulated part, and $g$ an optional target or dependency. Pick(bowl)
and Pick(cube) share an operation while keeping different object arguments.

Table~\ref{tab:atomic_skills} describes the operations in the included
simulation tasks. The prompts are examples from the demonstration conversion
scripts. They specify a behavior, rather than a fixed trajectory: the policy
still has to use the current scene and arm state to execute it.

\begin{table}[tbp]
\centering
\small
\renewcommand{\arraystretch}{1.12}
\caption{Atomic operations and example prompts. Object and target arguments
vary across tasks. These semantic names do not specify numerical adapter IDs.}
\label{tab:atomic_skills}
\begin{tabularx}{\linewidth}{@{}l>{\raggedright\arraybackslash}X>{\raggedright\arraybackslash}p{0.35\linewidth}@{}}
\toprule
Skill & Behavior and arguments & Example atomic prompt \\
\midrule
Pick & Approach and grasp the specified object. & Pick up the bowl. \\
Place & Carry a held object to the target and release it. & Place the can into the basket. \\
Lift & Raise a grasped object or container to the required pose. & Lift the basket. \\
Pull & Move a grasped articulated part to open it. & Pull the drawer open. \\
Move & Position the end effector for the next manipulation. & Move to the blue cube. \\
Push & Move an object toward a target through contact. & Push the blue cube to the target. \\
Recover & Return the arm after its manipulation phase. & Recover. \\
Wait & Maintain the current configuration until the other arm or a dependency allows progress. & Wait for the other robots. \\
\bottomrule
\end{tabularx}
\end{table}

\paragraph{Manipulation and waiting.}
Wait can mean remaining still with an empty gripper or continuing to hold
an object. In Can in Basket, the basket arm waits while the other arm places
the can. In Object in Cabinet, the drawer arm waits after opening the
drawer. Recover instead requests an arm motion after manipulation.
Neither word implies a new task goal. In the Push Cubes conversion, recovery
motion is included in the Push segment rather than given a separate Recover
prompt. Thus, a skill label describes the demonstrated phase, not every
motion that occurs within it.

\subsection{Skill Coverage Across Tasks}
\label{appendix:skill_coverage}

Table~\ref{tab:skill_coverage} connects the atomic operations to the six
source families and two held-out cross-task families. Reorder, Sync, and
Sync+Reorder keep a family's manipulation operations and change their
relations across arms. Cross-task composition combines source behaviors
with new partners or targets. Sharing an operation name alone does not
establish that an entire manipulation is familiar: the object, target, and
partner arm also determine what changes at test time.

\begin{table}[tbp]
\centering
\small
\renewcommand{\arraystretch}{1.10}
\caption{Manipulation operations and their roles in ACG-Bench. Source rows
summarize demonstration prompts; the final two rows describe held-out
compositions. Wait and recovery behavior are discussed in
Appendix~\ref{appendix:atomic_skills}.}
\label{tab:skill_coverage}
\begin{tabularx}{\linewidth}{@{}l>{\raggedright\arraybackslash}p{0.23\linewidth}>{\raggedright\arraybackslash}X@{}}
\toprule
Task family & Manipulation operations & Source behavior or held-out combination \\
\midrule
Stack Bowls & Pick, Place & Each arm handles a bowl and places it at its target. \\
Stack Cubes & Pick, Place & Each arm handles a cube and places it at its target. \\
Push Cubes & Move, Push & Each arm approaches a cube and pushes it to a target. \\
Burger \& Fries & Pick, Place & The arms place the burger and fries at their respective targets. \\
Can in Basket & Pick, Lift, Place & One arm holds and lifts the basket; the other places the can inside. \\
Object in Cabinet & Handle grasp, Pull, Pick, Place & One arm grasps and opens the drawer; the other places the box inside. \\
\midrule
Cube in Bowl & Pick, Place & Combine bowl handling and cube placement with a bowl target. \\
Bowl in Cabinet & Handle grasp, Pull, Pick, Place & Combine drawer opening with bowl placement. \\
\bottomrule
\end{tabularx}
\end{table}

The cabinet prompt ``Grab the drawer handle'' specifies the grasp that
precedes Pull. Its wording is retained here to distinguish handle grasping
from picking up a free object. The skill-label file determines how prompt
variants map to the ten adapter indices used by SkillLoRA
(Appendix~\ref{appendix:skill_lora}).

\subsection{From Demonstrations to Per-arm Prompts}
\label{appendix:phase_annotation}

The source trajectories are segmented before training. For Stack Bowls,
Stack Cubes, and Burger \& Fries, the conversion scripts use each arm's
gripper closing and opening events to separate picking from placement.
They then assign a pair of atomic prompts to each segment.
Table~\ref{tab:phase_prompts} gives the Stack Cubes example. A segment can
contain two active behaviors, such as the left arm recovering while the
right arm picks. Its phase index identifies the prompt pair; it is not a
single atomic-skill ID.

\begin{table}[tbp]
\centering
\small
\renewcommand{\arraystretch}{1.10}
\caption{The five prompt pairs in the Stack Cubes source conversion.
The two prompt strings are stored together and separated by arm at policy
input. This is a training annotation, not a fixed test-time clock.}
\label{tab:phase_prompts}
\begin{tabularx}{\linewidth}{@{}c>{\raggedright\arraybackslash}X>{\raggedright\arraybackslash}X@{}}
\toprule
Phase & Left-arm prompt & Right-arm prompt \\
\midrule
1 & Pick up the blue cube. & Wait for the other robots. \\
2 & Put the blue cube on the target. & Wait for the other robots. \\
3 & Recover. & Pick up the green cube. \\
4 & Wait for the other robots. & Place the green cube on the target. \\
5 & Wait for the other robots. & Recover. \\
\bottomrule
\end{tabularx}
\end{table}

Gripper transitions do not identify every behavior. Push Cubes keeps its
grippers closed, so its segmentation uses end-effector height and changes
in joint motion to separate approach, pushing, and the handoff between arms.
The cabinet script uses handle-grasp and joint-motion cues to locate the end
of pulling. These offline boundaries label the demonstrations; at evaluation,
the shared scheduler uses the task's state checks to update prompts.
The evaluator separately checks the required physical milestones and final
goal. A gripper closure in a demonstration is therefore an annotation cue,
not sufficient evidence of a successful test-time grasp.

\subsection{Examples of Unseen Compositions}
\label{appendix:composition_examples}

Consider a source routine in which the left arm picks and places its object
before the right arm does the same. Write $k_L,k_R$ for the two pick
milestones and $d_L,d_R$ for the two placement milestones. The source
order is $k_L\prec d_L\prec k_R\prec d_R$, where $\prec$ denotes a
required predecessor. Recovery and waiting prompts are omitted from this
example so that the manipulation events are easy to compare.

\begin{table}[tbp]
\centering
\small
\renewcommand{\arraystretch}{1.12}
\caption{How a familiar routine can be recomposed. The first three rows
illustrate changes in order and timing; the last illustrates source-skill
reuse across tasks. Exact milestones remain task-specific.}
\label{tab:composition_examples}
\begin{tabularx}{\linewidth}{@{}l>{\raggedright\arraybackslash}X@{}}
\toprule
Test group & Change to the source routine \\
\midrule
Reorder & Require $k_R\prec d_R\prec k_L\prec d_L$: the right arm completes its manipulation first. \\
Sync & Require $|\tau(k_L)-\tau(k_R)|\leq\delta$, followed by the source placement order $d_L\prec d_R$. \\
Sync+Reorder & Use the same synchronized pick pair, then reverse the placement order to $d_R\prec d_L$. \\
Cross-task & Combine bowl handling from Stack Bowls with cube placement from Stack Cubes; the cube's target is now a bowl. \\
\bottomrule
\end{tabularx}
\end{table}

Here $\tau(e)$ is the recorded completion time of event $e$, and $\delta$
is the task's synchronization window. Synchronization is checked on these
event times; it does not require the arms to have identical trajectories or
continuously overlap their motions. A cross-task condition remains in the
Cross-task group even when it also synchronizes events. Thus, Cube in
Bowl/Sync contributes once to the 17-condition generalization mean.

\section{Policy Implementation}
\label{appendix:implementation}

\subsection{Training Configuration}
\label{appendix:training_detail}

Table~\ref{tab:training_parameters} lists the multitask AE-VLA training
configuration. The four grouped variants turn SkillLoRA and AWA on or off.
AWA enables both arm-wise attention switches together. Token Group + AWA uses the multi-agent attention model without
skill banks. All four use the same source dataset, batch size, training
steps, and learning-rate settings.

\begin{table}[t]
\centering
\small
\caption{Multitask AE-VLA training configuration.}
\label{tab:training_parameters}
\begin{tabular}{ll}
\toprule
Component & Setting \\
\midrule
Initialization & $\pi_{0.5}$ base checkpoint \\
Backbone / action expert variants & \texttt{gemma\_2b} / \texttt{gemma\_300m} \\
Control dimensions / model dimensions & $2\times 8$ / 32 (zero-padded) \\
Action horizon / grouped action tokens & 50 / $2\times 50$ \\
Prompt/state token budget & 200 per arm; 400 in total \\
Optimizer & AdamW; $(\beta_1,\beta_2)=(0.9,0.95)$ \\
Optimizer $\epsilon$ / weight decay & $10^{-8}$ / $10^{-10}$ \\
Global gradient-norm clipping & 1.0 \\
Learning-rate warmup & 1,000 steps \\
Peak / final learning rate & $5\times10^{-5}$ / $5\times10^{-5}$ \\
Configured cosine schedule horizon & 200,000 steps \\
Training steps / global batch size & 30,000 / 32 \\
Checkpoint interval & 30,000 steps \\
Frozen parameters & None \\
EMA decay & 0.99 \\
Image-mask / wait-augmentation probability & 0 / 0 \\
Arm-shuffle probability & 0 \\
Latent collaboration tokens & Disabled \\
\bottomrule
\end{tabular}
\end{table}

Because the cosine schedule has equal peak and final rates, the learning rate
is constant after warmup. Joint actions are offsets from the
current joint state; gripper actions remain absolute. These settings disable extra image masking,
added Wait examples, and shuffling of arm roles. This does not disable the backbone's standard
training image preprocessing.
Single $\pi_{0.5}$ and every AE-VLA variant produce both arms from one
model instance. Dual $\pi_{0.5}$ trains and runs an independent instance
for each arm.

\paragraph{Flow-matching action loss.}
Let $a$ be a demonstrated joint action chunk, $o$ the policy inputs, and
$\epsilon$ Gaussian noise. At noise level $t$, training uses
\begin{equation}
    a_t=(1-t)a+t\epsilon,\qquad
    \mathcal{L}_{\mathrm{action}}=
    \mathbb{E}\!\left[\|v_\theta(a_t,t\mid o)-(\epsilon-a)\|^2\right].
    \label{eq:action_loss}
\end{equation}
The expectation samples a demonstration, a noise vector
$\epsilon\sim\mathcal{N}(0,I)$, and $t\sim\mathrm{Beta}(1.5,1)$ scaled
to $[0.001,1]$. The squared error is averaged over action dimensions and
chunk positions. The joint prediction contains both arms' velocities.
At inference, ten Euler steps denoise the joint action chunk.

\subsection{Shared Instruction Interface and SkillLoRA}
\label{appendix:skill_lora}

Single $\pi_{0.5}$, MA-VLA, Dual $\pi_{0.5}$, and every AE-VLA variant use
the same per-arm atomic prompts at evaluation. Each prompt identifies the
active skill and its object, target, or dependency arguments. The scheduler
uses the same phases and state checks for every method;
the time at which a prompt changes can differ because the policies produce
different trajectories.

\paragraph{Skill labels and routing.}
The ten adapter IDs are atomic skill types, not full tasks or skill--object
pairs. During training, the data loader maps dataset task IDs to per-arm
skill IDs using \texttt{skill\_labels.jsonl}. These labels train the routers.
All methods receive the same skill information in their evaluation prompts,
but SkillLoRA adds a skill classification loss during training.

The configuration enables \texttt{router\_use\_text\_only} and
\texttt{router\_per\_arm}. Each arm has its own two-layer MLP
($2048\!\rightarrow\!1024\!\rightarrow\!10$, with a Swish hidden activation).
The router averages its arm's prompt/state features after prefix attention,
ignoring padding. The text-only option selects the positions to average;
those features include discrete state tokens and have already attended to
allowed image tokens. The router can therefore use visual and state information.
It selects the highest-scoring skill and encodes that choice as a one-hot
vector with no gradient through the choice. Training and inference both use
this prediction. Ground-truth labels are used only in the training loss. One selection per arm is reused
for all 50 action tokens and all ten Euler denoising steps.

\paragraph{Adapted layers and sharing.}
Each adapted layer has ten pairs of rank-4 factors,
with scale $16/4=4$. The banks are applied to every one of the
action expert's 18 transformer layers: query, key/value, and attention-output
projections, both feed-forward gating branches, and the feed-forward output
projection. The action input and output projections also use skill banks.
The visual encoder and vision--language prefix do not use skill banks.
Adapters are shared across arms; router MLPs are arm-specific.
The shared model weights and the added parameters are jointly trained.
For a matrix with $d_{\mathrm{in}}$ inputs and $d_{\mathrm{out}}$ outputs,
the ten rank-4 adapters add $40(d_{\mathrm{in}}+d_{\mathrm{out}})$ parameters.
This expression applies to each adapted attention projection and feed-forward
branch.

\paragraph{Loss and interpretation.}
The router loss is cross-entropy averaged over valid arm labels, with weight
0.1; missing labels are represented by $-1$ and excluded. The hard choice blocks action-loss gradients through the router.
The classification loss can also update the prefix features used by the
router. SkillLoRA changes the action layers, number of weights, and training
loss. A shared adapter with the same parameter count or a shuffled-label
control could help separate these effects. We test execution with supplied
skill prompts; we do not test incorrect instructions or skill labels.

\subsection{Token Layout and Exact Attention Masks}
\label{appendix:attention_detail}

Fixed token ranges and output assignments identify each arm.
There are no learned arm-ID embeddings. Each arm's instruction
is tokenized together with its own discretized state. The $\pi_{0.5}$
suffix contains two action-token blocks and no separate continuous-state
tokens. Both blocks embed the full noisy joint action vector, and their
decoded outputs are concatenated into the joint prediction. Thus, the mask restricts attention connections,
but it does not make the two action streams independent.

\paragraph{From joint tokens to arm-specific tokens.}
In Single $\pi_{0.5}$, each action token represents one step of the joint
action: decoding that token produces both arms' outputs. Its suffix has
$H=50$ tokens. Token Group instead uses $[A_L,A_R]$, with $H$ tokens
in each block, for $2H=100$ action tokens in total. The left block supplies
the left-arm output and the right block supplies the right-arm output at
each horizon position. Both blocks pass through the same action expert and
output projection. Their outputs are assembled into one joint velocity
chunk, which is integrated during denoising.

For simulation, the noisy input and velocity prediction both have shape
$H\times32$: 16 physical control dimensions and 16 padding dimensions.
Grouping changes the action-token sequence length, not the control horizon
or number of controlled joints. The action input projection receives the
full 32-dimensional noisy vector in both groups. The per-arm organization
therefore refers to token ranges and output assignment, not to separate
noisy inputs or independent policy instances.

\paragraph{Why grouping supports the two design choices.}
The prefix layout is $[G,W_L,W_R,P_L,P_R]$, where each $P_r$ contains
the arm's instruction and discretized state. The local group
$C_r=W_r\cup P_r$ links that observation to action block $A_r$.
SkillLoRA applies the selected update $\Delta W_{\hat{s}_r}$ to the
tokens in $A_r$. AWA uses the same group boundaries to determine which
keys each query may read. Grouping supplies the arm assignment; the two
mechanisms separately change weights and attention connections.
All four combinations use this layout. The Token Group comparison with
Single $\pi_{0.5}$ also changes suffix length from $H$ to $2H$.

With AWA, global prefix tokens can attend to every valid prefix token.
Each arm's wrist and prompt/state tokens can attend to the global tokens
and their own local group. An action token can read the same prefix group
and its own arm's action tokens at the same or earlier chunk positions.
Prefix queries never attend to the action suffix.

Equation~\ref{eq:arm_attention} in the main text gives the complete AWA
key sets. Padding masks apply to all three cases.

The \texttt{full\_attn} variants disable both arm-wise switches and use
the standard block mask. Their prefix is bidirectional; action
queries can access every valid prefix token. Within the suffix, tokens
attend bidirectionally inside each arm block, and a later block can attend
to earlier blocks. With the stored $[A_L,A_R]$ order, $A_R$ can attend to
$A_L$, but $A_L$ cannot attend to $A_R$. With AWA, both cross-arm action
directions are blocked and each block becomes lower triangular.
The comparison therefore changes both cross-arm and within-chunk attention.
It tests the whole mask, rather than blocking cross-arm attention alone
while keeping the connections within each chunk fixed.

\subsection{Relationship to Prior Design Principles}
\label{appendix:design_correspondence}

Our quantitative study holds the pretrained backbone initialization fixed
at $\pi_{0.5}$. We adapt MA-VLA~\citep{zhang2026mavla} from its original
$\pi_0$ backbone to $\pi_{0.5}$, and the MA-VLA label throughout the paper
refers to this adaptation. TwinVLA uses different pretrained weights, so
including it would mix differences in pretraining with differences in
architecture. We therefore omit its results from the quantitative
comparison. Below, we clarify architectural differences from prior work.

TwinVLA~\citep{im2026twinvla} starts from its pretrained single-arm
SingleVLA and copies the VLM into two arm-specific branches. The branches
share the visual encoder and a diffusion-transformer action head, which
jointly decodes their readout tokens. Joint attention combines queries,
keys, and values from both branches, applies a mask permitting cross-arm
access, and splits the outputs back into the branches. The projection and
feed-forward weights remain arm-specific.

TwinVLA also processes language and ego-view tokens as a shared sequence.
A learned gate mixes the two arms' feed-forward experts for these tokens;
attention reweighting preserves the contribution of shared modalities.
Our Token Group keeps global context and separate arm token ranges in
one $\pi_{0.5}$ backbone, without duplicating the VLM. Its two action-token
groups create places to apply arm-specific routing and attention rules.
SkillLoRA selects atomic-skill adapters on action tokens, rather than
mixing arm experts on shared observation tokens. AWA blocks direct
cross-arm action attention while retaining a path through global context,
whereas TwinVLA's joint attention permits direct exchange between branches.
These variants therefore test related design choices within $\pi_{0.5}$;
they are not a reproduction of TwinVLA's full architecture.

SkillVLA~\citep{zhai2026skillvla} studies skill reuse and communication
between arms. Our variants test per-arm token groups, skill-specific action
weights, and attention access within one backbone. They do not include
SkillVLA's high-level skill selector or adaptive cooperation gate.
A released SkillVLA implementation was unavailable for these experiments.
The study compares design ideas under the same training and evaluation
setup; it does not rank the full TwinVLA and SkillVLA systems.

\section{Additional Results and Evaluation Protocol}
\label{appendix:evaluation}

\subsection{Task Results}
\label{appendix:task_results}

Table~\ref{tab:macg_task_results} lists all 23 conditions for the three
baseline policies and AE-VLA.

\subsection{Complete Architectural Results}
\label{appendix:architecture_results}

Table~\ref{tab:architecture_tasks} reports the full $2\times2$ comparison
on Token Group. Token Group + AWA was scored from its 2,300 episode records
using the same set of conditions, task-specific synchronization windows,
and scoring rules as the other variants. Each condition uses seeds
1000--1099.

\begin{table}[t]
\centering
\small
\setlength{\tabcolsep}{4pt}
\caption{\textbf{Four design settings on all conditions: CCSR (\%).} TG means Token Group; SL means SkillLoRA. Each setting uses one policy and the same 23 conditions, with 100 episodes per condition.}
\label{tab:architecture_tasks}
\begin{tabular}{llrrrr}
\toprule
Task & Condition & TG & TG + SL & TG + AWA & AE-VLA \\
\midrule
Stack Bowls & In-domain & \textbf{88.00} & 83.00 & 43.00 & 83.00 \\
 & Reorder & 0.00 & 32.00 & 0.00 & \textbf{79.00} \\
 & Sync & 0.00 & 19.00 & 0.00 & \textbf{26.00} \\
 & Sync+Reorder & 0.00 & 17.00 & 7.00 & \textbf{67.00} \\
\midrule
Stack Cubes & In-domain & \textbf{34.00} & \textbf{34.00} & 8.00 & 20.00 \\
 & Reorder & 0.00 & 3.00 & 1.00 & \textbf{19.00} \\
 & Sync & 0.00 & 0.00 & 1.00 & \textbf{3.00} \\
 & Sync+Reorder & 0.00 & 0.00 & 0.00 & \textbf{21.00} \\
\midrule
Push Cubes & In-domain & 3.00 & \textbf{21.00} & 17.00 & 8.00 \\
 & Reorder & 0.00 & 0.00 & \textbf{6.00} & 3.00 \\
 & Sync & \textbf{4.00} & 0.00 & 1.00 & \textbf{4.00} \\
 & Sync+Reorder & \textbf{1.00} & 0.00 & 0.00 & 0.00 \\
\midrule
Burger \& Fries & In-domain & \textbf{20.00} & 10.00 & 1.00 & 6.00 \\
 & Reorder & 0.00 & 0.00 & 0.00 & \textbf{5.00} \\
 & Sync & 0.00 & \textbf{2.00} & 0.00 & 1.00 \\
 & Sync+Reorder & 0.00 & 0.00 & \textbf{1.00} & 0.00 \\
\midrule
Can in Basket & In-domain & \textbf{74.00} & 49.00 & 40.00 & 25.00 \\
 & Sync & 1.00 & 6.00 & \textbf{21.00} & 18.00 \\
\midrule
Cube in Bowl & Cross-task (Reference) & 0.00 & 2.00 & 3.00 & \textbf{54.00} \\
 & Cross-task (Sync) & 1.00 & 0.00 & 0.00 & \textbf{11.00} \\
\midrule
Object in Cabinet & In-domain & \textbf{51.00} & \textbf{51.00} & 41.00 & 21.00 \\
 & Sync & 0.00 & 0.00 & \textbf{2.00} & 0.00 \\
\midrule
Bowl in Cabinet & Cross-task (Reference) & \textbf{58.00} & 21.00 & 56.00 & 55.00 \\
\bottomrule
\end{tabular}
\end{table}

\subsection{Real-world SO101 Setting}
\label{appendix:so101}

\begin{figure*}[htbp]
    \centering
    \includegraphics[width=\linewidth]{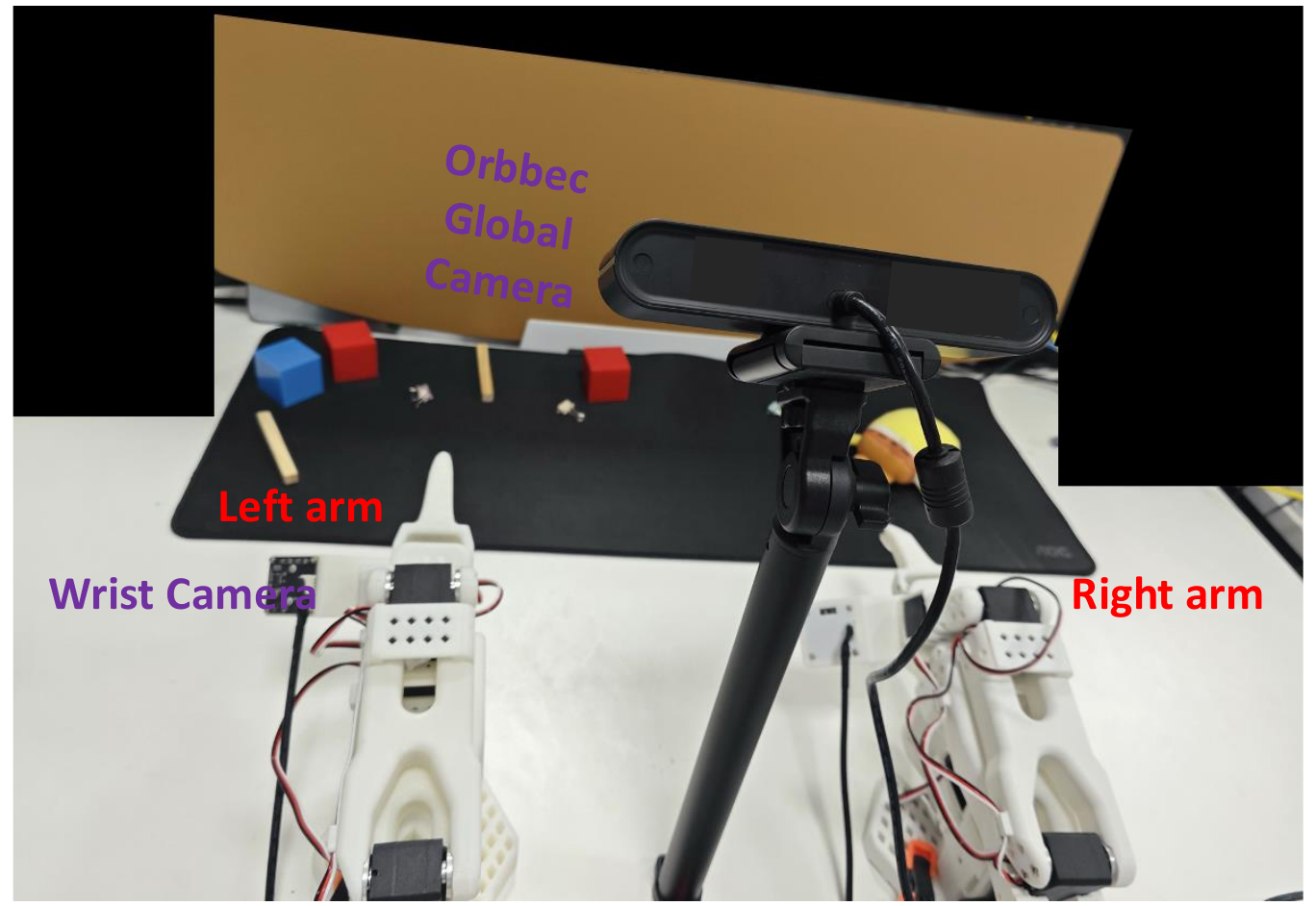}
    \caption{\textbf{SO101 setup.} Two arms share a workspace observed by
    one fixed global camera and one wrist camera on each arm.}
    \label{fig:real_robot_setup}
\end{figure*}

\begin{figure*}[htbp]
    \centering
    \includegraphics[width=\linewidth]{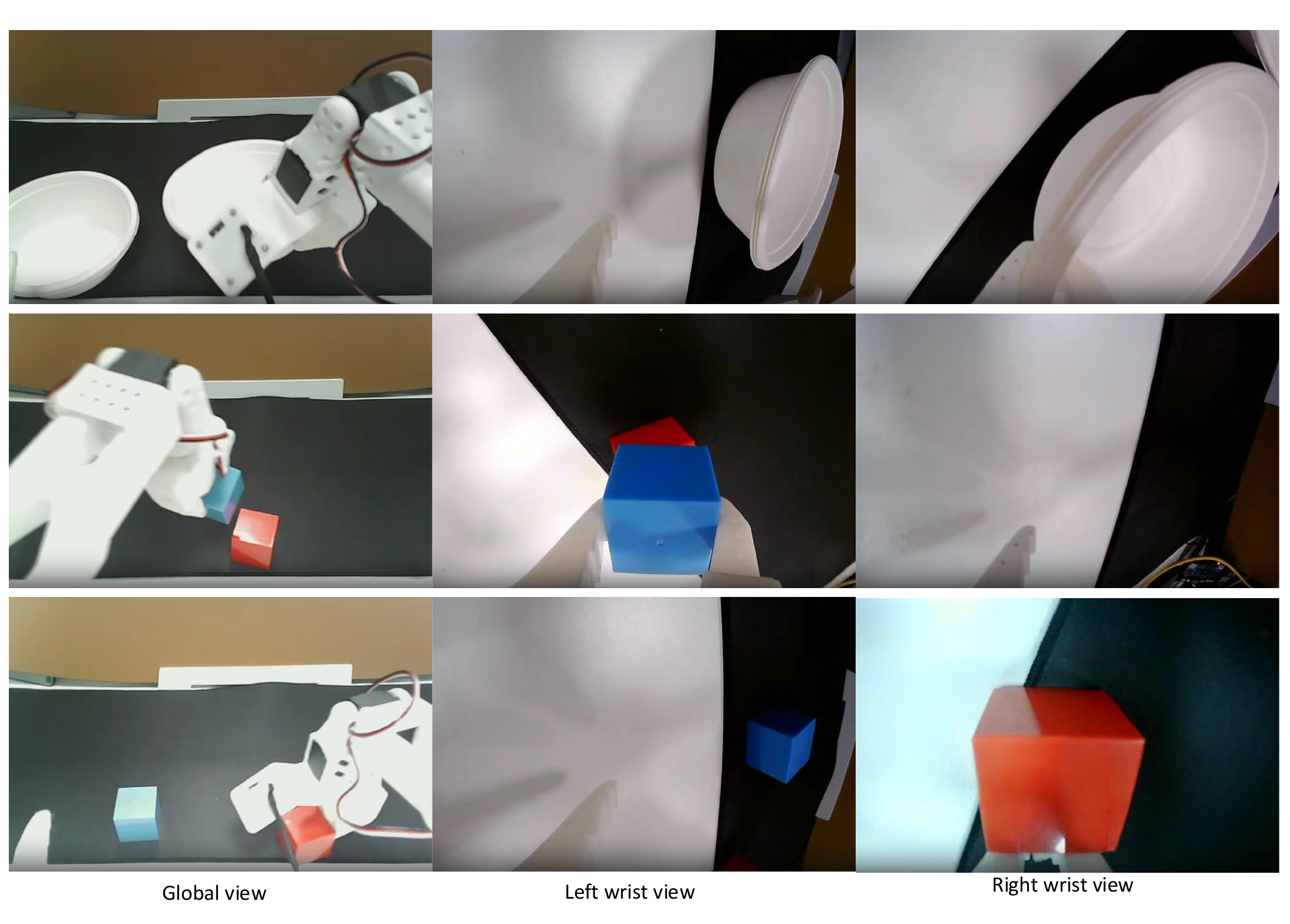}
    \caption{\textbf{SO101 training demonstrations.} Example observations
    from the global, left-wrist, and right-wrist cameras (left to right).
    We collect 50 demonstrations per training task through LeRobot teleoperation.}
    \label{fig:real_robot_training}
\end{figure*}

Figure~\ref{fig:real_robot_setup} shows the real-world platform, which
consists of two SO101 arms with 6 degrees of freedom
per arm. One fixed RGB camera observes the shared workspace and one wrist RGB
camera is attached to each arm; all three streams run at 30\,Hz.

We collect 50 demonstrations for each of Stack Bowls, Stack Cubes, and
Push Cubes through LeRobot teleoperation~\cite{cadene2024lerobot}, giving
150 demonstrations in total. Figure~\ref{fig:real_robot_training} shows
example camera observations. We annotate the demonstrations with the same
arm-wise phase vocabulary used in simulation and pool all three tasks
when training each real-world policy. The training hyperparameters match
those used for the simulation benchmark; the grouped-policy configuration
is given in Appendix~\ref{appendix:training_detail}. No Cube in Bowl
demonstrations are used for training.
The cube pick-and-place behavior is learned from Stack Cubes; placing the cube
in a bowl is a held-out composition evaluated only at test time.

We evaluate eight conditions with 20 rollouts each. The three In-domain
conditions are Stack Bowls, Stack Cubes, and Push Cubes. The generalization
conditions are Reorder and Sync variants of Stack Bowls and Push Cubes, plus
Cube in Bowl/Cross-task Combination. Each rollout randomizes valid object poses and
contains distractor objects. In-domain success requires the task goal and all
physical milestones. Generalization success additionally requires the target
order or synchronization relation; Cross-task Combination requires completion
of the recombined bowl-and-cube phase graph. The table reports success counts
for individual conditions and mean rates across the three In-domain
and five generalization conditions. Each method has 160 evaluated rollouts,
for 640 rollouts in total.

\subsection{Constraint-compliant Scoring}
\label{appendix:scoring}

For conditions without synchronization, we check milestones in the required
order. An episode succeeds only if it reaches the final environment goal
and completes all milestones under the task rules. For Sync and
Sync+Reorder, the evaluator records the first unconditional completion time of
each milestone. Both designated synchronized milestones must occur, with a time gap no
larger than the task window. Later milestones must follow this pair in the
required order; tied completion times are allowed. Normalized progress is
the number of milestones completed under these rules divided by the number
required.

Can in Basket is an exception: in-domain success uses the environment's
\texttt{task\_success} directly. Its Sync condition also requires the two
pick events to occur within 20\,s. Object in Cabinet/Sync uses a 15\,s
window; all other included Sync and Sync+Reorder conditions use 45\,s.
These checks concern event times, not the overlap of whole action intervals.

Manual review found arm collisions in all apparent Dual $\pi_{0.5}$ successes
on Stack Bowls/Sync and Cube in Bowl/Sync. Their CCSR is set to zero while
their milestone progress is retained. Progress is therefore not a measure
of collision-free execution.

The Single $\pi_{0.5}$ and MA-VLA logs were generated before the final
synchronization windows were fixed, so we rescored them from their recorded
milestone timelines. As a consistency check, the same scorer reproduces all
cached condition-level flags and progress values for the other five variants,
except for the two documented manual collision overrides. Single $\pi_{0.5}$
and MA-VLA each have 2,300 evaluated episodes, with seeds 1000--1099 in every
condition.

\end{document}